\documentclass[sigconf,screen,nonacm]{acmart}

\usepackage[T1]{fontenc}
\usepackage{array}
\usepackage{graphicx}
\usepackage{fancyvrb,xstring}
\usepackage{subcaption}
\usepackage{pythonhighlight}
\usepackage{listings}
\usepackage{multirow}
\usepackage{multicol}
\usepackage{booktabs}
\usepackage{xspace}
\usepackage{bm}
\usepackage{pifont} 

\usepackage{xcolor,tikz}

\usepackage{amsmath,amsfonts,bm}

\def\eqref#1{equation~\ref{#1}}

\def\1{\bm{1}}

\DeclareMathAlphabet{\mathsfit}{\encodingdefault}{\sfdefault}{m}{sl}
\SetMathAlphabet{\mathsfit}{bold}{\encodingdefault}{\sfdefault}{bx}{n}

\DeclareMathOperator*{\argmin}{arg\,min}

\newcommand{\fig}[1]{Fig.~\ref{#1}}

\newcommand{\tb}[1]{Tab.~\ref{#1}}
\newcommand{\se}[1]{Section~\ref{#1}}

\newcommand{\cmark}{\ding{51}}%
\newcommand{\xmark}{\ding{55}}%

\usepackage{color}
\usepackage{xcolor}

\newcommand{\our}{\textsc{RITA}\xspace}
\newcommand{\ourbk}{\textsc{RITABackend}\xspace}
\newcommand{\ourkit}{\textsc{RITAKit}\xspace}

\AtBeginDocument{%
  \providecommand\BibTeX{{%
    \normalfont B\kern-0.5em{\scshape i\kern-0.25em b}\kern-0.8em\TeX}}}

\setcopyright{acmcopyright}
\copyrightyear{2023}
\acmYear{2023}
\setcopyright{acmlicensed}\acmConference[DAI '23]{The Fifth International Conference on Distributed Artificial Intelligence}{November 30-December 3, 2023}{Singapore, Singapore}
\acmBooktitle{The Fifth International Conference on Distributed Artificial Intelligence (DAI '23), November 30-December 3, 2023, Singapore, Singapore}
\acmPrice{15.00}
\acmDOI{10.1145/3627676.3627681}
\acmISBN{979-8-4007-0848-0/23/11}

\begin{document}

\title{
RITA: Boost Driving Simulators \\with Realistic Interactive Traffic Flow
}

\author{Zhengbang Zhu$^*$\textsuperscript{1$\dagger$}, Shenyu Zhang$^*$\textsuperscript{1}, Yuzheng Zhuang$^*$\textsuperscript{2}, \\
Yuecheng Liu\textsuperscript{2}, Minghuan Liu\textsuperscript{1}, Ziqin Gong\textsuperscript{1}, Shixiong Kai\textsuperscript{2}, Qiang Gu\textsuperscript{2}, \\
Bin Wang\textsuperscript{2}, Siyuan Cheng\textsuperscript{2}, Xinyu Wang\textsuperscript{3}, Jianye Hao\textsuperscript{2}, Yong Yu\textsuperscript{1}
}
\affiliation{%
\institution{\textsuperscript{\rm 1}Shanghai Jiao Tong University, \textsuperscript{\rm 2}Noah's Ark Lab, Huawei Technologies Co., Ltd., \\
\textsuperscript{\rm 3} IAS BU ADS Planning and Control, Huawei Technologies Co., Ltd.
\country{}
}
\textsuperscript{$\dagger$}zhengbangzhu@sjtu.edu.cn
}

\renewcommand{\shortauthors}{Z. Zhu et al.}

\begin{abstract}

High-quality traffic flow generation is the core module in building simulators for autonomous driving.
However, the majority of available simulators are incapable of replicating traffic patterns that accurately reflect the various features of real-world data while also simulating human-like reactive responses to the tested autopilot driving strategies.
Taking one step forward to addressing such a problem, we propose \textbf{R}ealistic \textbf{I}nteractive \textbf{T}r\textbf{A}ffic flow (\our) as an integrated component of existing driving simulators to provide high-quality traffic flow for the evaluation and optimization of the tested driving strategies.
RITA is developed with consideration of three key features, i.e., fidelity, diversity, and controllability, and consists of two core modules called \ourbk and \ourkit. \ourbk is built to support vehicle-wise control and provide traffic generation models from real-world datasets, while \ourkit is developed with easy-to-use interfaces for controllable traffic generation via \ourbk.
We demonstrate RITA's capacity to create diversified and high-fidelity traffic simulations in several highly interactive highway scenarios. The experimental findings demonstrate that our produced RITA traffic flows exhibit all three key features, hence enhancing the completeness of driving strategy evaluation.
Moreover, we showcase the possibility for further improvement of baseline strategies through online fine-tuning with RITA traffic flows.

\end{abstract}

\ccsdesc[500]{Computing methodologies~Distributed artificial intelligence}
\ccsdesc[500]{Computing methodologies~Inverse reinforcement learning}




\maketitle
\def\thefootnote{*}\footnotetext{The first three authors contributed equally to this work.}\def\thefootnote{\arabic{footnote}}

\section{INTRODUCTION}
With the empowered ability of artificial intelligence, autonomous driving has become one of the promising directions for both research and application.
Due to the safety concern and the high cost of real vehicle testing, autopilot simulation is widely used for fast iteration and verification of driving algorithms~\cite{dosovitskiy2017carla, rong2020lgsvl}.
It attracts massive attention from researchers and industries to build a high-quality simulator.
However, existing open-sourced autopilot simulators primarily focus on traffic modeling with hand-scripted rules to facilitate lane following, collision avoidance, etc. ~\cite{elsayed2020ultra, krajzewicz2002sumo}. Such heuristic rules are insufficient to model the variability and noise in real-world human driving behaviors.


To improve the fidelity of driving behavior simulation, replaying the movement in the human driving dataset~\cite{osinski2020carla} has become popular.
However, the data replay lacks fidelity during real-time interactions, since the social vehicles cannot provide reasonable responses when the ego vehicle acts differently as in the dataset.
To further generate both realistic and reactive traffic flows, supervised learning approaches have been applied in social vehicle modeling~\cite{bansal2018chauffeurnet, bergamini2021simnet}.
Unfortunately, supervised decision models suffer from compounding errors~\cite{xu2020error} and causal confusion~\cite{de2019causal} problems, which make them perform poorly as the planning steps and traffic density increase. Adversarial imitation learning methods handle these issues through rich online interactions in driving simulators~\cite{bhattacharyya2018multi, bhattacharyya2019simulating}. However, few works have explored the potential of using them to enhance the traffic flow in existing simulators.

\begin{figure*}[ht]
\centering
\includegraphics[width=0.95\linewidth]{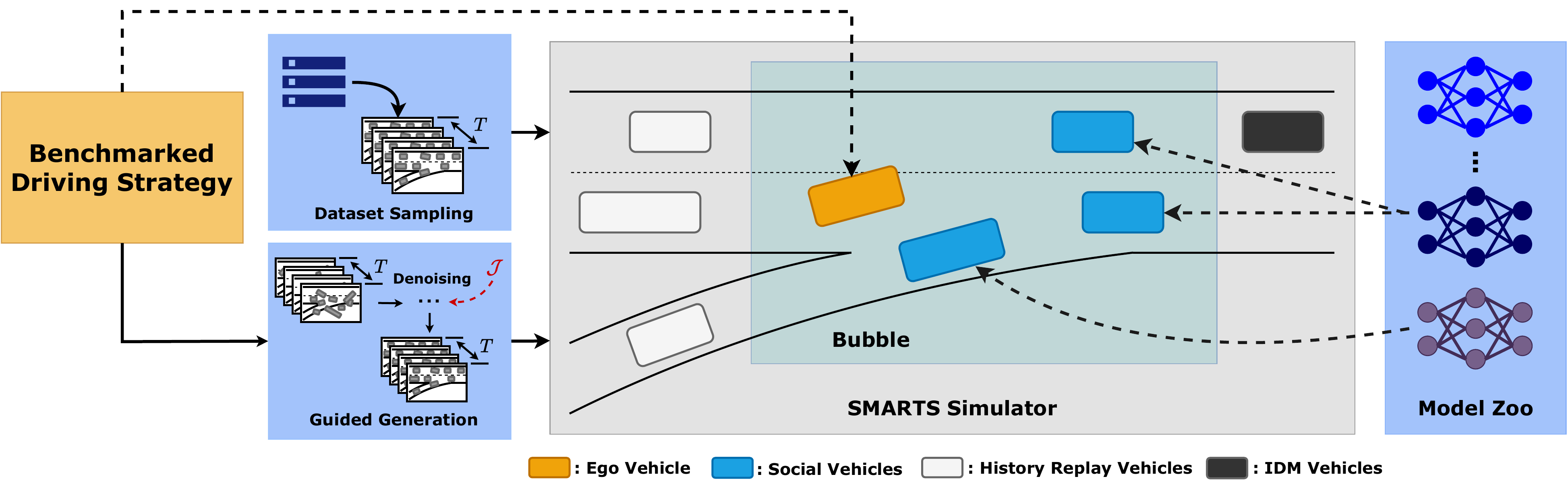}
\vspace{-5pt}
\caption{Overview of the \our traffic generation pipeline.}
\vspace{-10pt}
\label{fig:overview}
\end{figure*}

Another advantage of using simulations is that it allows users to design extensive and specific scenarios beyond those from the datasets to evaluate driving strategies.
Prior learning-based works in building traffic flow only use a single model or multiple models separately~\cite{zhou2020smarts, kothari2021drivergym}, which limits the configurability of the simulator.
A recent work, SimNet~\cite{bergamini2021simnet}, replaces the model-inferred steering angle with hard-coded values to force generating lane-changing behaviors of social vehicles. This approach sacrifices fidelity to provide specific interaction scenarios.
Instead, our focus is to enable specific scenario generations while maintaining fidelity.

To provide a principled and unified solution for building high-quality traffic flows in driving simulators, in this paper, we propose \our, a traffic generation framework with \textbf{R}ealistic \textbf{I}nteractive \textbf{T}r\textbf{A}ffic flow.
\our focuses on augmenting existing autonomous driving simulators with high-quality traffic flows, aiming at reaching three design goals: \textit{fidelity}, \textit{diversity}, and \textit{controllability}. To meet these requirements, \our is developed with two core modules: \ourbk and \ourkit.
The first module, \ourbk, incorporates machine learning models built from real-world datasets, including vehicle control and guided traffic generation models.
In particular, the vehicle control models are trained using adversarial imitation learning methods, which generate specific behaviors and generalize better outside the data distribution than supervised methods. In addition, the guided traffic generation models are for customized background traffic generation, providing specific initialization states with a guided diffusion sampling algorithm.
The second module, \ourkit, is an easy-to-use toolkit that combines models from \ourbk to generate traffic flows in the simulator. \ourkit has a friendly programming interface that makes it easy to set up the different traffic flows that can be controlled.

To show the superiority of generated traffic flows and demonstrate the usages of \our, we conduct experiments on two class of representative tasks in highway driving, \textit{cut-in} and \textit{ramp}. With regard to these tasks, we first assess the quality of the generated traffic flow with regard to three design goals. Then we show two use cases of \our to facilitate the autonomous driving workflow. On the one hand, \our traffic flow can be used flexibly to benchmark off-the-shelf driving strategies from different aspects. Benchmarked driving strategies are either hand-crafted or trained with the history-replay traffic flow. Also, \our traffic flow can help further improve the driving strategies' performances by allowing online interactions with reactive and realistic social vehicles.
Currently, \our is in the process of being incorporated into Huawei's autonomous driving platform.
In the future, we hope \our can serve as a standard building block in developing high-fidelity autonomous driving simulators and contribute to the evaluation and improvement of driving strategies for the community.

\section{Key Features}

We highlight three key features of the traffic flow in RITA.

\paragraph{Fidelity}
Fidelity is an essential measure of simulators that describes the gap between the simulation and the reality \cite{zhang2021learning}.
In particular, driving simulators are essential for ensuring the sim-to-real performance of driving algorithms and are the most effective evaluation tool before real-world road testing. Thus, the smaller the gap between the interaction responses in the simulator and those encountered in reality, the more likely it is that a strategy achieving high performance in the simulator will also perform well in real road conditions.
The fidelity of the traffic flow lies in two aspects: 1) the vehicle state distribution in the traffic flow. For example, vehicles always move fast on highways in light traffic conditions; 2) the fidelity of the responses from vehicles from the traffic flow when interacting with other vehicles, like the typical lane-changing scenario. We observe that human drivers exhibit specific noise in multiple dimensions when performing lane changes in real datasets. Therefore, we would expect that the traffic flows generated by \our also possess similar microscopic properties to human interaction behaviors.

\paragraph{Diversity}
Human driving behavior exhibits a high degree of diversity due to differences in destinations, road conditions, drivers' personalities, and other factors. This means, in similar situations, three distinct drivers will behave differently: if driver A tries to change lanes to the left when there is a vehicle in his front, driver B can tend to maintain his current route; even if driver C also chooses to change lanes, the timing of action and the speed of lane change can be quite different. 
To reach that, we must ensure that \our's basic vehicle control models include as much behavioral diversity as possible from real data. In addition, when these models are combined to generate diversified traffic flow, the resulting behavior still maintains a high fidelity level. 

\paragraph{Controllability}
Building upon diversity, we can define controllability as an enhancement feature, i.e., the ability to generate customized traffic flows based on user specifications from many potential instances. For example, we can construct test tasks with different difficulty levels by adjusting the distribution and behavior of the traffic flow, or we can generate scenarios with a targeted assessment of the weaknesses of the driving strategy under test.
High controllability also corresponds to high usability, allowing various researchers to build their benchmark tasks by simply changing the parameters of the traffic definition interface in \ourkit. The modular design of \our allows it to be easily extended to more maps and datasets, or even used by other simulators.

\section{\our Overview}

In this section, we present an overview of our proposed traffic generation framework.
To achieve the three design goals, the framework needs to be able to have sufficient expressiveness to cover the diverse traffic flows in the real data and enough variable modules to support the generation of specific traffic flows based on the users' specifications. At a glance, we maintain a diverse zoo of vehicle control models and multiple static replay trajectories generation methods, and propose a set of interfaces for traffic definition to achieve the above requirements.

We start with some necessary definitions, and then give a compositional illustration of \our.

\subsection{Definitions}
The goal of this paper is to build microscopic traffic flows that can interact with user-specified driving strategies, which are referred to as the \textit{ego strategies} in the following text.
Thereafter, we divide the vehicles as \textit{ego vehicles} and \textit{social vehicles}. Ego vehicles are those that are controlled by ego strategies, while social vehicles' movements are determined by the simulator, data replay, or the trained reactive agents as in \se{sec:backend}. Note that such divisions do not have to be fixed during simulation. For example, one vehicle can be a social vehicle in the first half of an episode and is then taken over by the ego strategies for the second half.

\subsection{Compositional Illustration}
\our's aspiration is to create high-quality traffic flow to improve the existing driving simulators. Developing a simulator and building basic components (such as kinematic simulation, collision detection, etc.) from scratch are outside the major scope of concern. To this end, we choose to build upon an existing simulator, in this paper particularly, SMARTS~\cite{zhou2020smarts}, yet it is feasible to integrate \our into any other simulators such as SUMO~\cite{krajzewicz2002sumo} and BARK~\cite{bernhard2020bark}.

\our can be decomposed into two core modules, \ourbk and \ourkit. As a high-fidelity traffic generation framework, \our can be initialized from real-world datasets and transform traffic specifications to specific scenarios.

\begin{figure}[h]
\centering
\includegraphics[width=0.98\linewidth]{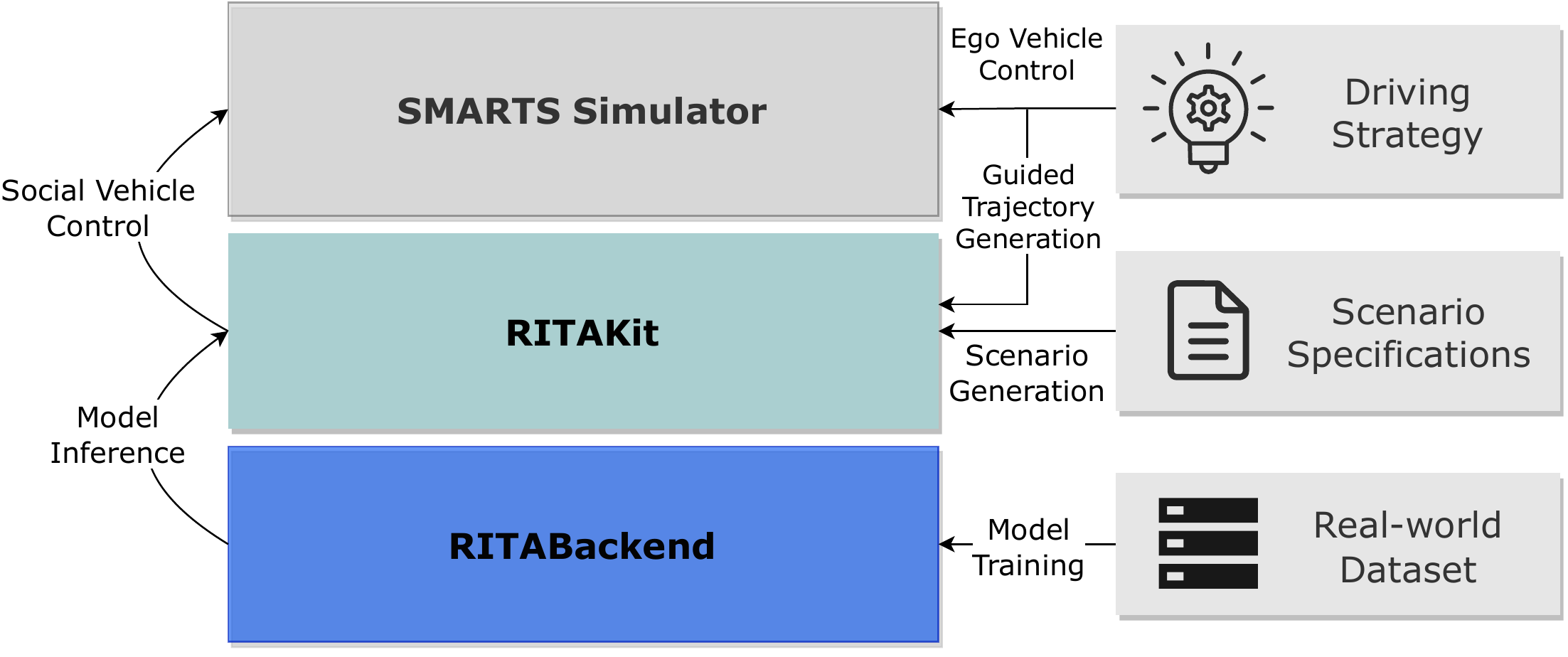}
\caption{Compositional illustration of \our.}
\label{fig:components}
\end{figure}

We illustrate the architecture of \our in \fig{fig:components}, organized by a bottom-to-top order. First, real-world datasets are used to train machine learning models in \ourbk. Once models are built, the scenarios can be configured via \ourkit with scenario specifications, usually expressed as parameters in a unified traffic generation function.
Notably, \ourkit directly operates on \ourbk and transforms scenario specifications into model calls in \ourbk. With this decoupled design, the creation and adjustment of benchmark tasks can be made with less effort, where users do not need to define each sub-model call manually.
During the evaluation procedure, both \ourkit and ego strategies are plugged into SMARTS to control social vehicles and ego vehicles, respectively. 
\ourkit also takes ego strategies as input for rare-case generation and automatically generates customized traffic flows.

In the following, We begin by introducing four example real-world tasks used to demonstrate the construction and use of the \our in \se{sec:tasks}. After that, we show the training and evaluation results for models in \ourbk in \se{sec:backend}, and describe \ourkit in detail in \se{sec:toolkit}. In \se{sec:benchmark}, we demonstrate the benchmarking results of different ego strategies and further optimization of these strategies on \our traffic flows.


\section{Evaluated Tasks}
\label{sec:tasks}
We focus on four typical interactive tasks on highway driving to showcase the complete pipeline of building \ourbk and using \our to benchmark ego strategies via \ourkit. We construct these tasks from the open-source NGSIM I80 dataset for training and evaluation. 

\emph{Task 1: Left Cut-in scenario.} 
Social vehicles in the lane to the right of the lane in which ego vehicles are located will attempt to change lanes to the left.

\emph{Task 2: Right Cut-in scenario.} 
It is a symmetrical situation with Task 1, when the social vehicles try to change to the right lane where ego vehicles are located.

\emph{Task 3: On-ramp scenario.} 
This task handles the classic ramp road structure in the highway scenario.
Ego vehicles start from the right-most lane of the highway road, and there is a constant flow of social vehicles merging in from the ramp on the right.

\emph{Task 4: Off-ramp scenario.} 
In this benchmarking task, ego vehicles also start from the right-most lane. Social vehicles surrounding ego vehicles attempt to exit the highway from the right-hand ramp.

In this paper, the training of models in \ourbk, the scenario specifications for \ourkit, and the evaluation and optimization of ego strategies are based on these four tasks. 
We provide schematic illustrations of these four tasks in Appendix \se{sec:task-illustration}. 


\section{Building \ourbk}
\label{sec:backend}

\subsection{Model Zoo}
To evaluate the ego strategies with high-fidelity traffic flows, we build a set of reactive agents that can generate traffic flows with flexible interactions. 
Instead of designing complex and proper reward functions for reinforcement learning driver agents, our reactive agents learn to behave like human drivers by imitating real-world data. To this end, we utilize imitation learning methods~\cite{ho2016generative,liu2020energy}, which provide a way to learn from demonstrations and mimic the demonstrator's behaviors. In \our, we maintain a model zoo that integrates a set of imitation learning algorithms for building reactive agents.
There exist some previous methods that take behavior cloning (BC) to learn their reactive agent models, such as~\cite{9561666,bansal2018chauffeurnet}, due to its simplicity. However, BC can suffer from serious compounding error problem when data coverage is limited~\cite{ross2010efficient}. 
On the contrary, inverse reinforcement learning (IRL) methods~\cite{arora2021survey} theoretically outperform BC with less compounding error~\cite{xu2020error}.
Thereafter, we take off-the-shelf Generative Adversarial Imitation Learning (GAIL) algorithm~\cite{ho2016generative} and its variants to build agent models in our model zoo. 
In the following, we briefly introduce these methods and their features for driving strategy learning.

\begin{figure*}[ht]
    \centering
    \begin{subfigure}[b]{0.23\textwidth}
        \centering
        \includegraphics[width=\textwidth]{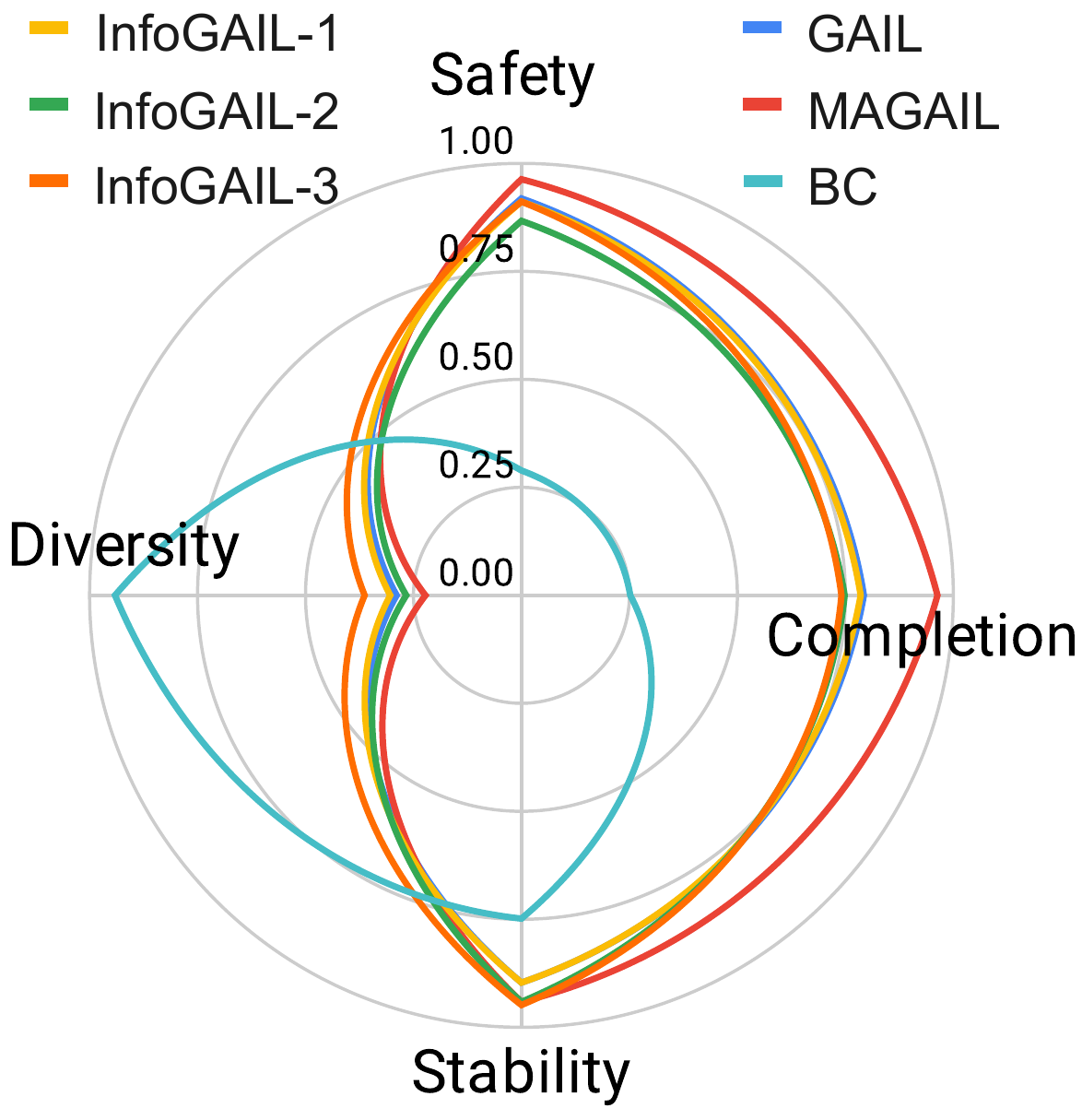}
        \caption{Left Cut-in}
    \end{subfigure}
    \begin{subfigure}[b]{0.23\textwidth}
        \centering
        \includegraphics[width=\textwidth]{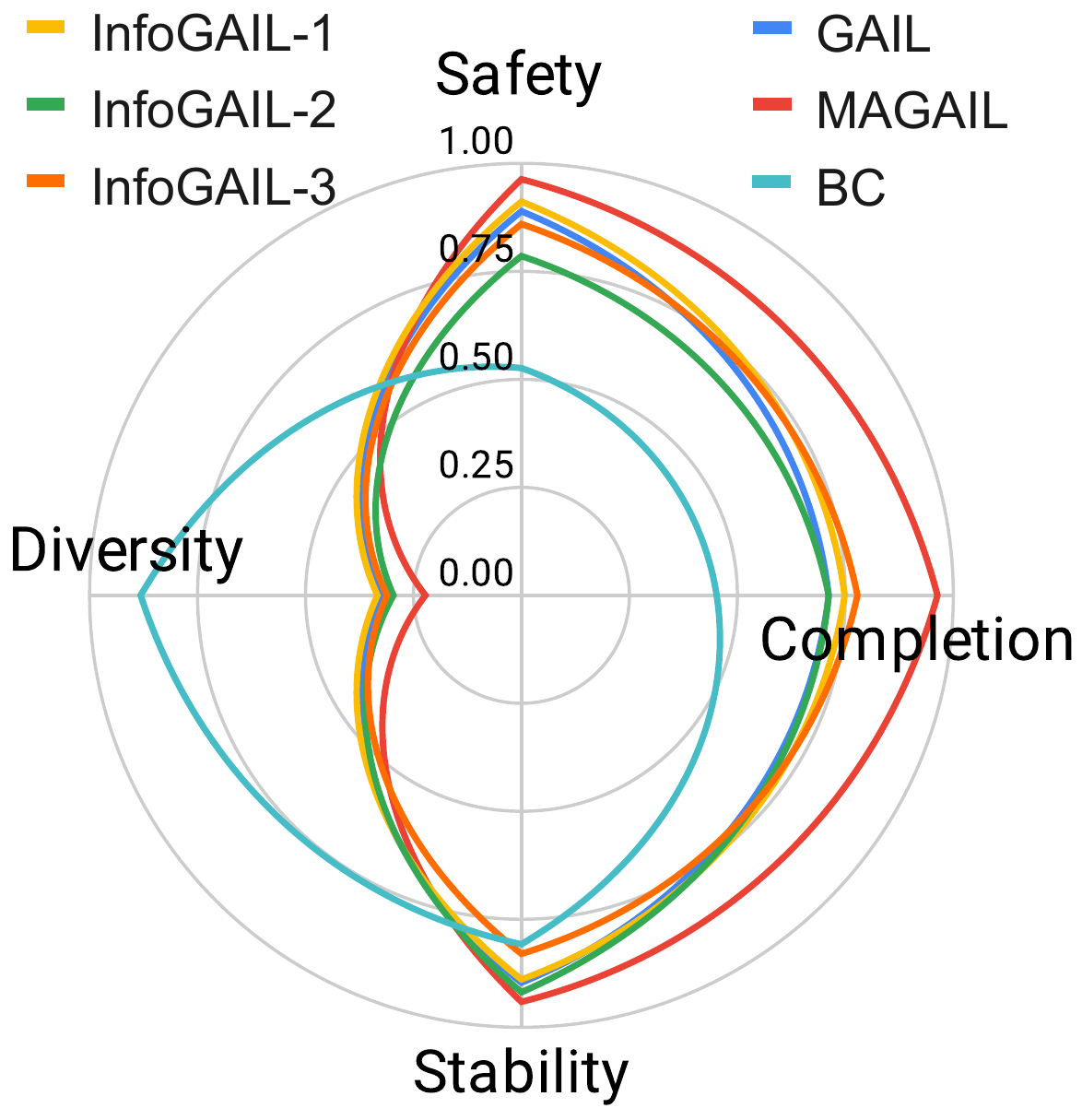}
        \caption{Right Cut-in}
     \end{subfigure}
    \begin{subfigure}[b]{0.23\textwidth}
        \centering
        \includegraphics[width=\textwidth]{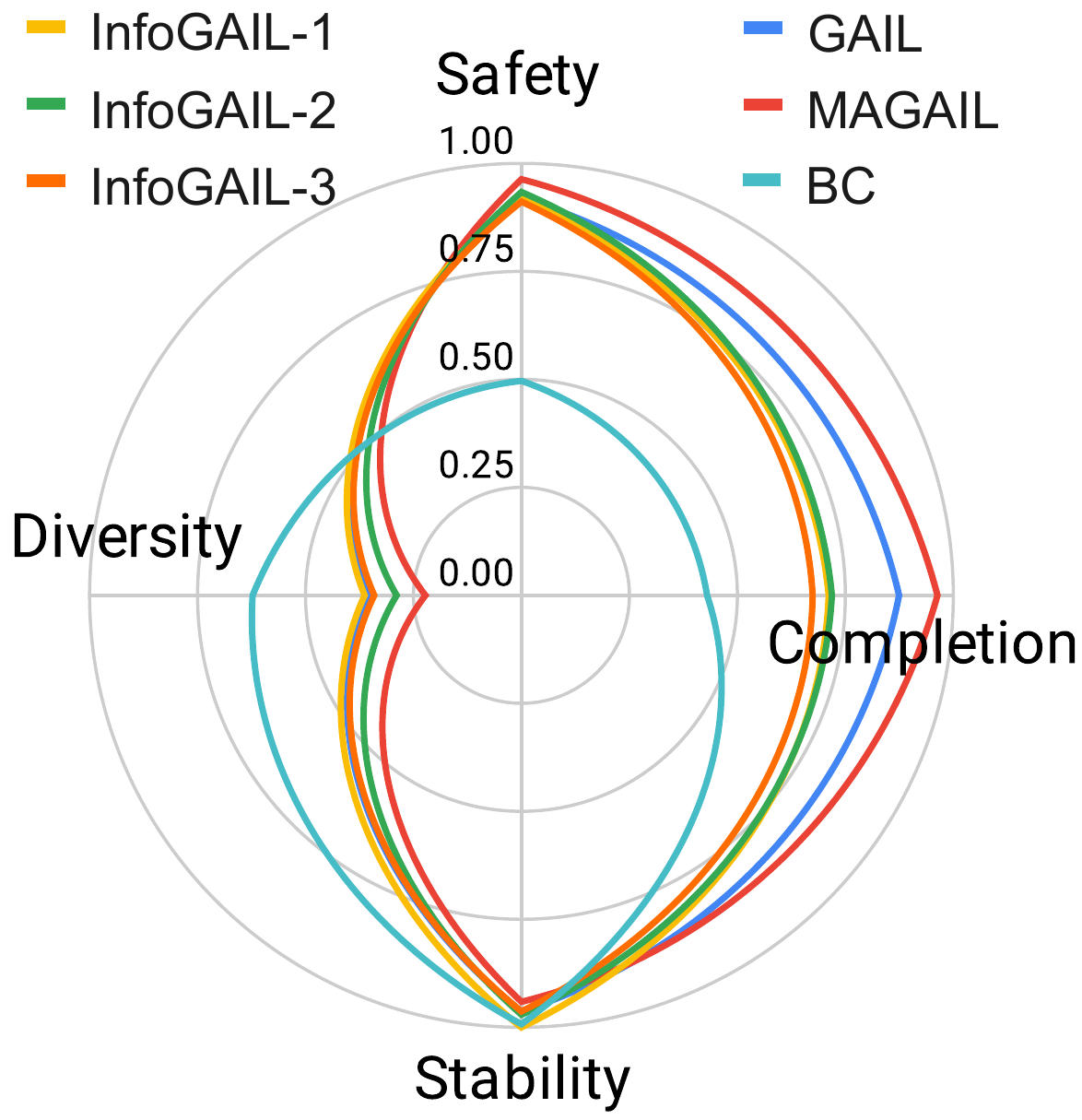}
        \caption{On-ramp}
     \end{subfigure}
    \begin{subfigure}[b]{0.23\textwidth}
        \centering
        \includegraphics[width=\textwidth]{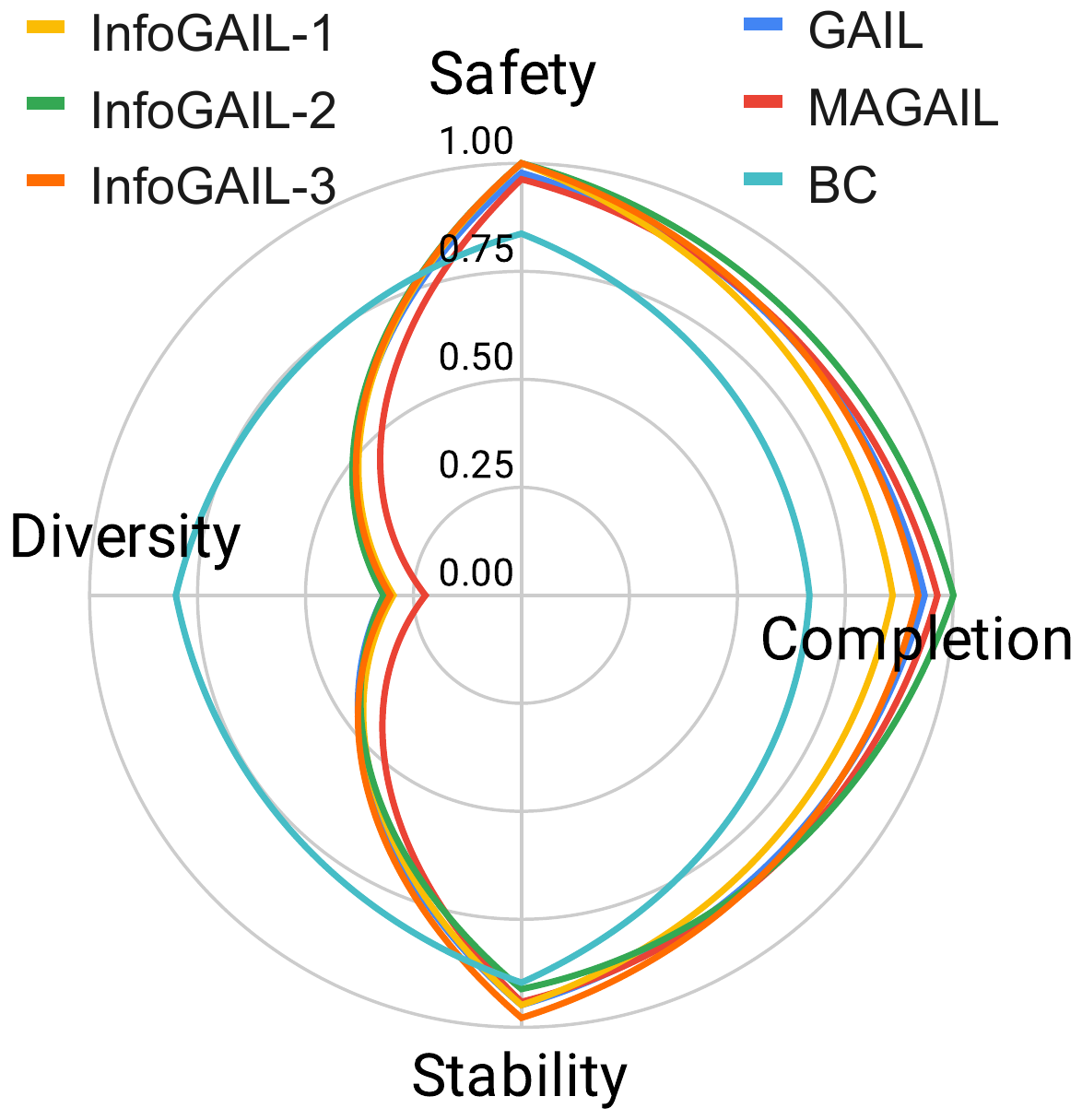}
        \caption{Off-ramp}
     \end{subfigure}
    \caption{Performance of vehicle control models in the model zoo. We evaluate the performance from four aspects, where \textit{Safety} and \textit{Completion} focus on the driving task itself; \textit{Stability} and \textit{Diversity} measure the statistical characteristics of agent actions in driving trajectories.}
\vspace{-5pt}
\label{fig:Model Performance}
\end{figure*}

\subsubsection{Single-agent GAIL}
We apply GAIL, a single-agent algorithm, to multi-agent learning problems by independently learning different agents' strategies.
GAIL learn each agent's policy by training a discriminator that distinguishes expert data from the one sampled by the learned policies.
Formally, GAIL optimizes the following objective:
\begin{equation*}
\min\limits_{\pi}  \max\limits_{D} \mathbb{E}_{\pi_E} [\log D(s,a)] + \mathbb{E}_{\pi} [\log(1 -  D(s,a))] - \lambda H(\pi)
\end{equation*}
where $D$ is the discriminator that distinguishes expert state-action pairs from policy-generated pairs, $H$ is the casual entropy term, and $\lambda$ is the hyperparameter.
In our implementation, we train the policy model by iteratively replacing one history vehicle in the replay data as the ego vehicle to be controlled.
Although GAIL can not automatically detect and extract different behavior modes, individual GAIL models can exhibit distinct and desired reactive behaviors if they are trained to imitate a pre-selected portion of demonstration data (e.g., left cut-in).

\subsubsection{InfoGAIL}
We include InfoGAIL~\cite{li2017infogail} in our model zoo to capture the diversity of human driving patterns and produce high-quality interactive scenarios. The latent variable in InfoGAIL models can automatically detect and represent different distinct modalities of the unlabeled recorded data.
Since InfoGAIL is also a single-agent algorithm, the implementation follows the independent training style as GAIL. \our can provide diverse scenario configurations by controlling the latent variable.

\subsubsection{MAGAIL}
Considering the multi-agent nature of autonomous driving and modeling the interaction between different agents, we also include the multi-agent extension of GAIL, i.e., MAGAIL~\cite{song2018multi}, to serve as a multi-agent reactive agent. 
During training, a parameter-sharing MAGAIL model~\cite{8593758} controls all the vehicles in an area.
We want MAGAIL to model a general reactive policy that does not have to make any active interaction to influence the benchmarked strategies to be evaluated but instead makes passive responses only to ensure safe driving. 

\subsection{Model Performance Analysis}
\label{sec:model-performance}
To show the high fidelity of the learned models, we evaluate their performance from four aspects:
%
\begin{itemize}
    \item \emph{Safety}: Non-collision rate with social vehicles.
    \item \emph{Completion}: Completion rate of finishing specific interaction behavior. 
    \item \emph{Stability}: A constant minus the mean value of acceleration and yaw rate for the trajectory. Here, we select the constant as 1.0 for better visualization. 
    \item \emph{Diversity}: Standard deviation of acceleration and yaw rate for the trajectory.
\end{itemize}

We evaluate whether these models are capable of serving as social vehicle strategies. Therefore, we use the trained models to control one social vehicle, and other social vehicles and ego vehicles follow the recorded trajectories in the dataset. 
Since MAGAIL models are not required to complete specific interaction behaviors, their completion scores are set to the same as safety scores.
The results are shown in \fig{fig:Model Performance}. It is easily concluded that all GAIL-based models enjoy high safety and completion rate, which outperforms BC a lot, showing their capability for traffic generalization. The high stability provides smooth driving behavior similar to humans, along with some model-specific character. In particular, we find that the multi-agent reactive agent (MAGAIL) has the highest safety rate, while the multi-modal reactive agent (InfoGAIL) differs in diversity. These features indicate the necessity of multiple training algorithms for diverse models. Notably, the model performance does not seem high enough to create totally safe traffic, which may be because we enforce the model to make specific interactive behavior, resulting in the loss of reactivity. However, complete safety is meaningless since autonomous driving simulation needs exactly the realistic but unsafe environment for policy to make self-improvement.

\subsection{Guided Traffic Generation}
When utilizing \our to provide traffic flows for interacting with ego vehicles, a natural way is making all social vehicles controlled by reactive agents from the model zoo. However, such an approach can be computationally expensive when there exists numerous vehicles. Also, when an increasing number of vehicles react to each other, there is a growing chance of deviating from model training distributions. 
To save simulation cost and alleviate the aforementioned distributional shifts, we pre-specify a special region, which is called \textit{bubble}, on the map. Typically, the bubble covers the area inside which reactive behaviors are mostly expected to happen, such as the ramp merge area. Only social vehicles inside the bubble are controlled by reactive models, while others are just replaying movements in static trajectories.
In such a context, these static trajectories profoundly affect the interactions within the bubble by determining the initial states of social vehicles entering it. Even if we fix the control model of social vehicles within the bubble, the performance of ego vehicles under different static trajectories can vary much due to factors such as traffic density and the distance between social vehicles and ego vehicles.

Normally, the static trajectories can be sampled using \textit{dataset sampling}, i.e., directly sampled from datasets. Although dataset sampling maintains a high degree of fidelity, it has a slight chance to sample trajectories from the long tail part of its distribution (e.g., a group of quite dense traffic) as the dataset can be potentially large. Also, the sampled trajectories are agnostic to benchmarked strategies; therefore, dataset sampling overlooks the fact that every strategy has its Achilles heel.

\our provides \textit{guided generation} for acquiring static trajectories with more flexibility and controllability. First, a diffusion-based generative model $s_\theta$ is trained in \ourbk to cover the multi-vehicles trajectories distribution of the dataset.
The diffusion model is trained with the same objective as in DDPM, which is a reweighed version of the evidence lower bound (ELBO):
\begin{align*}
    \theta^*=\argmin_{\theta}&\sum_{i=1}^N(1-\alpha_i)\mathbb{E}_{p(x)}\mathbb{E}_{p_{\alpha_i(\tilde{x}|x)}}[\|s_{\theta}(\tilde{x}, i)\\
    &-\nabla_{\tilde{x}}\log p_{\alpha_i}(\tilde{x}|x)\|_2^2]~,
\end{align*}
where $p(x)$ is the multi-vehicle trajectories distribution in the dataset, $\alpha_i=\prod_{j=1}^i(1-\beta_j)$,  $p_{\alpha_i}(\tilde{x}|x)=\mathcal{N}(\tilde{x};\sqrt{\alpha_i}x, (1-\alpha_i)\bm{I})$, and $0<\beta_1,\beta_2,\cdots\beta_N<1$ is a sequence of positive noise scales. 
After training, users can specify a (differentiable) scoring function on trajectories, whose gradient is used for guided sampling from the diffusion model~\cite{song2020score}. A prediction network $\mathcal{J}(x)$ trained on a small labeled dataset is well suited as the scoring function. The generated static trajectories are guided towards the high-score instances while still within the real data support. 

Since it is difficult to define the order of vehicles in a multi-vehicle trajectory, the diffusion model uses a shared U-Net structure to process each vehicle's trajectory and perform an order-independent self-attention operation on all vehicles' intermediate features of the last encoder layer. The score function we used shares the same encoder structure with the diffusion model, but only performs attention operation between ego vehicle and other vehicles rather than a complete self-attention. This is because the metrics labels we used to train the scoring functions are primarily egocentric.



\section{\ourkit: A Configurable Traffic Generation Toolkit}
\label{sec:toolkit}
In this section, we present \ourkit, a traffic flow generation toolkit for interactive scenarios. \ourkit is a middle-ware that connects user specifications and \ourbk to generate traffic flow. From the user perspective, \ourkit is an easy-to-use programming interface. The following code block shows an example of creating a user-specific scenario.

\begin{figure}[h]
\centering
\includegraphics[width=\linewidth]{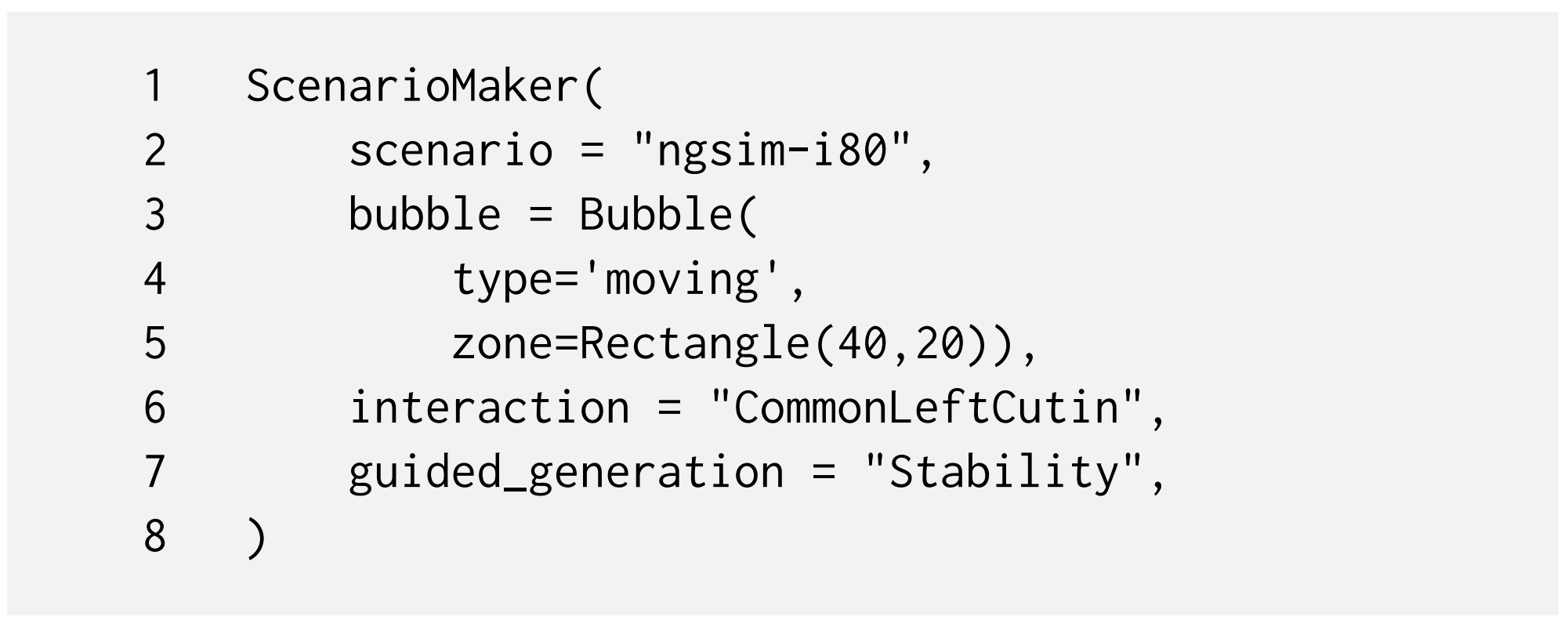}
\label{fig:ritakit}
\vspace{-10pt}
\end{figure}

We elaborate on each input parameter with how it influences the traffic flow and currently supported options.
\begin{itemize}
    \item \textit{scenario} specifies the map and the corresponding real-world dataset used to build models in \ourbk. Two scenarios, \texttt{ngsim-i80} and \texttt{ngsim-us101}, are already integrated and tested. We are continuously working to incorporate more open-source datasets.
    \item \textit{bubble} defines a special region in the roadway area to provide high quality and the most reactive traffic flow. Only social vehicles inside the bubble are controlled by the models from \ourbk. Before entering the bubble, the traffic is simply replaying the dataset, and after exiting the bubble, vehicles are controlled by IDM~\cite{treiber2000congested}. Bubble can be classified into two types, \texttt{moving} and \texttt{fixed}. The moving bubble is attached to and always covers the ego vehicle, while the fixed bubble is kept still in the specified positions.
    \item \textit{interaction} configures the distribution of models that control social vehicles inside the bubble.
    There are well-designed built-in options to choose from, e.g., CommonLeftCutin, CommonOnRamp, RareLeftCutin and RareOnRamp. In 'Common' interaction, models will be uniformly sampled from the model zoo, while 'Rare' interaction uses more proportion of InfoGAIL models. 
    \item \textit{guided\_generation} controls whether guided states generation is used and the score function for guiding. For example, \texttt{None} means not using guided generation but uniformly sampling from datasets instead. \texttt{Stability} refers to built-in score function stands for generating states that make the benchmarked strategy get worse stability, according to our designed metric. We support score function reconstruction in configurable interfaces as well.
\end{itemize}

We can use the above \ourkit interface to build the four tasks introduced in \se{sec:tasks}. For the two cut-in tasks, a moving bubble is set around the ego vehicle. And for on-ramp and off-ramp tasks, fixed bubbles are set in the surrounding area where the ramp and the highway road intersect.

\subsection{Generalized Traffic Flow Quality Measurement}
In this subsection, we evaluate the quality of traffic flows generated by \ourkit specifications. Different from \se{sec:model-performance} where individual models are evaluated, here we assess the performance of various models when they are used together to take over all social vehicles. In \fig{fig:Benchmark Quality Measurement}, we present results on left cut-in and on-ramp tasks, and results on the other two tasks can be found in the Appendix.

\subsubsection{Interaction Behavior Analysis}
\begin{figure*}[t]
\centering
\begin{subfigure}[b]{0.33\textwidth}
        \centering
        \includegraphics[width=\textwidth]{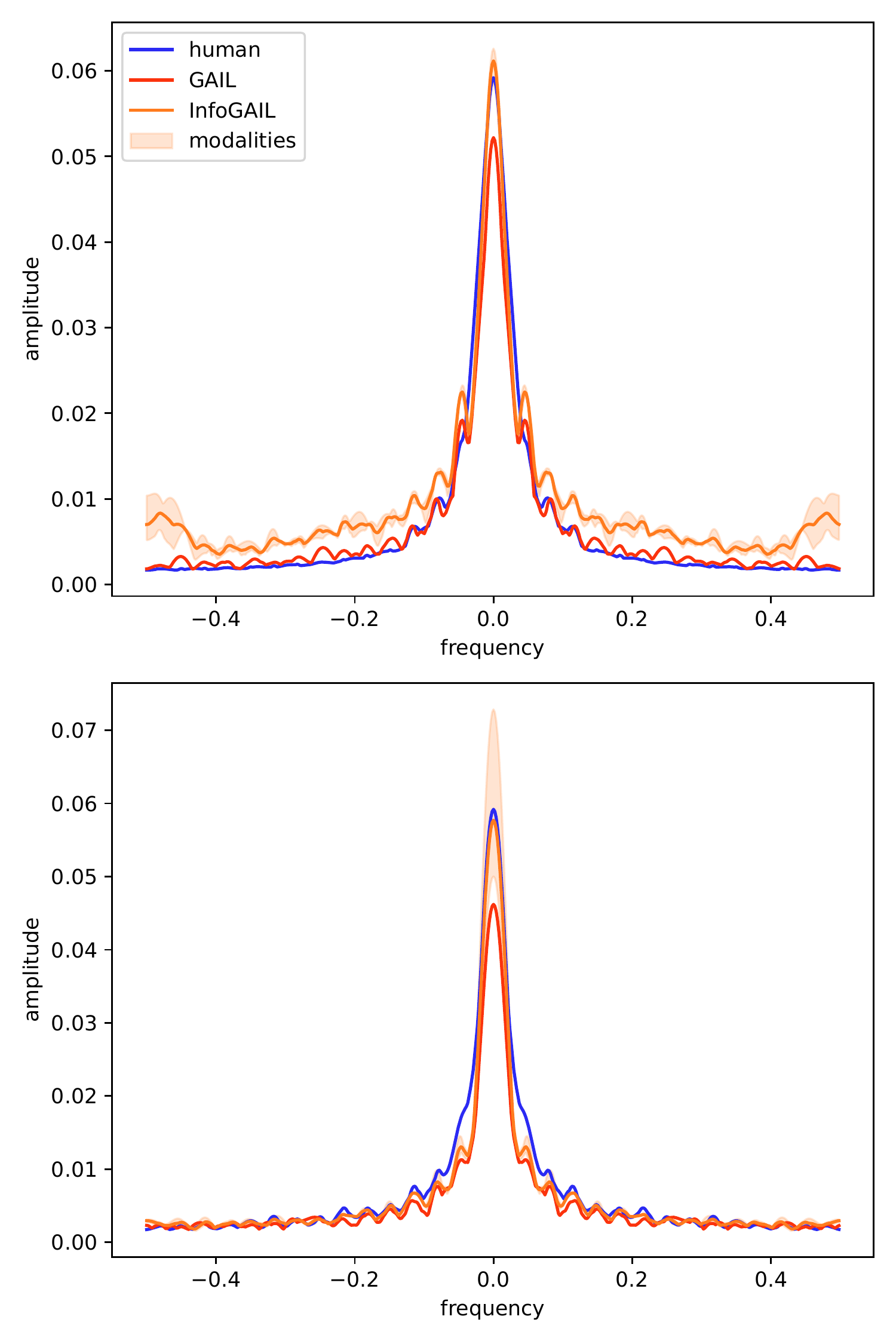}
        \caption{Fourier Analysis}
        \label{fig:Fourier Analysis}
\end{subfigure}
\begin{subfigure}[b]{0.33\textwidth}
        \centering
        \includegraphics[width=\textwidth]{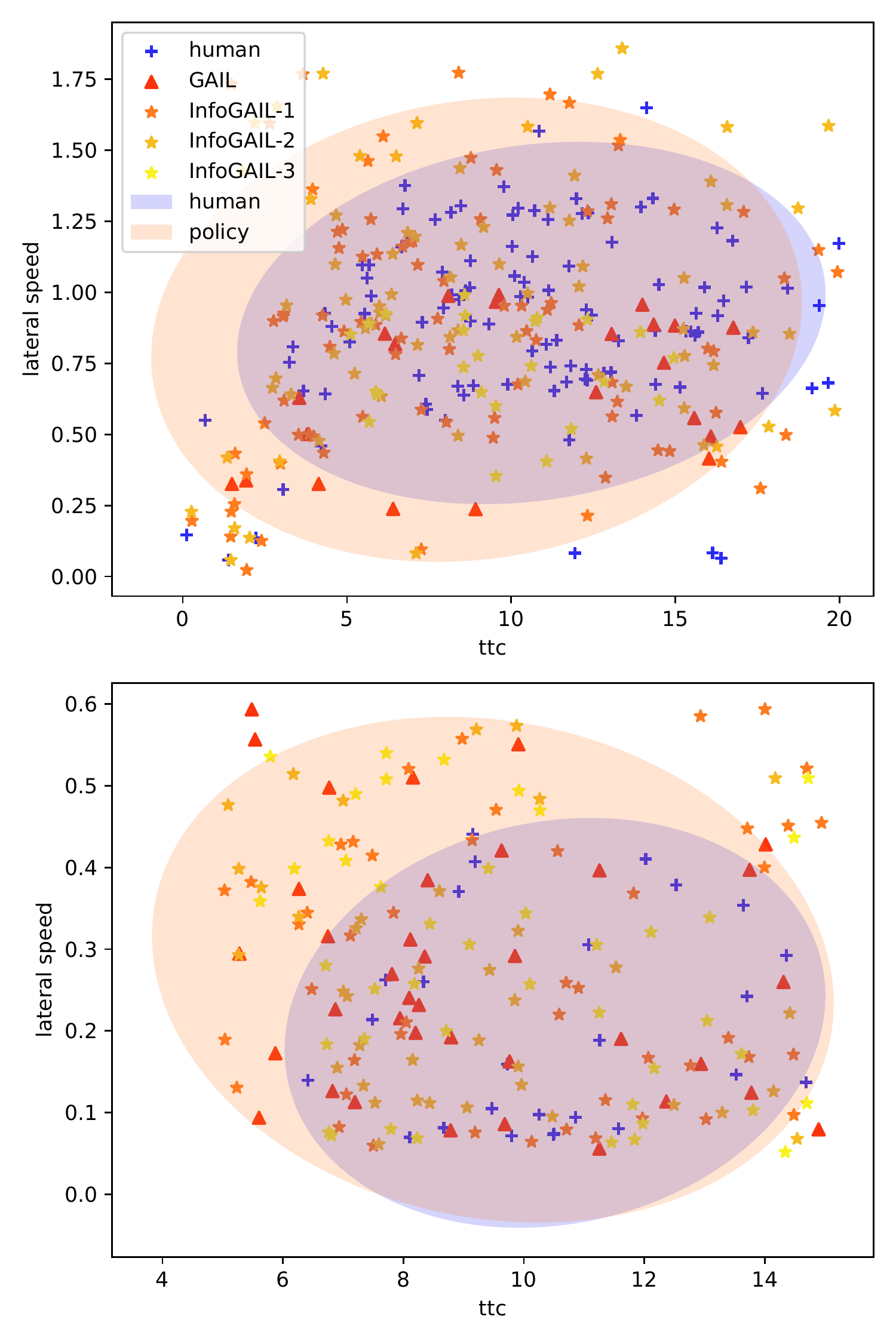}
        \caption{Scatter Distribution}
        \label{fig:Scatter Distribution}
\end{subfigure}
\begin{subfigure}[b]{0.267\textwidth}
        \centering
        \includegraphics[width=\textwidth]{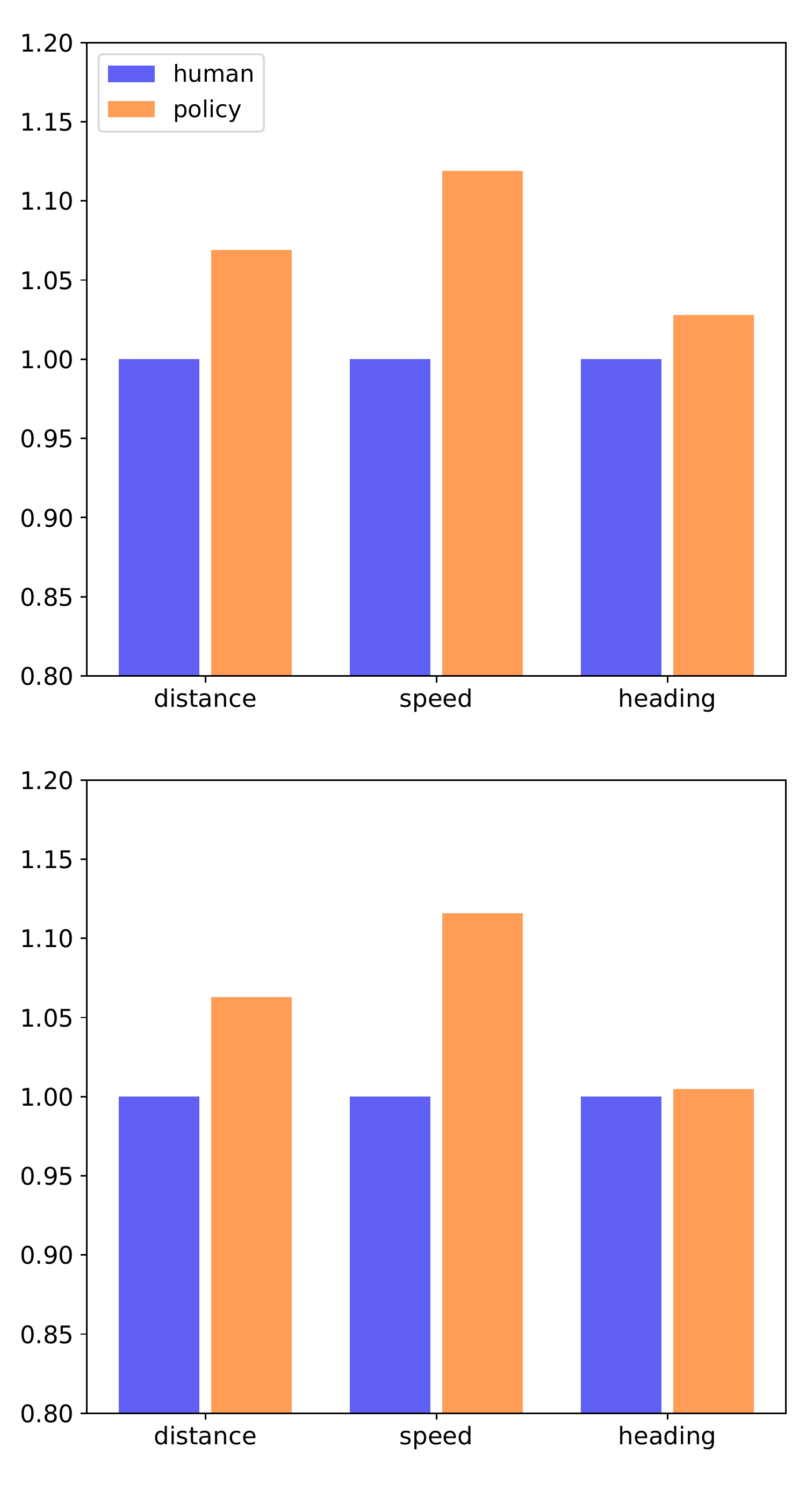}
        \caption{Traffic Flow Metrics}
        \label{fig:Traffic Flow}
\end{subfigure}
\caption{Measurements of traffic flow qualities. The first row corresponds to the left cut-in scenario, and the second to the on-ramp scenario. For interaction, (a) Fourier Analysis discusses lateral speed features over time, while (b) Scatter Distribution focuses on the cut-in moment when the most vital interaction happens. For traffic, (c) Traffic Flow Metrics quantifies the error of neighbors' statistics between human data and policy-generalized data in an egocentric view.}
\label{fig:Benchmark Quality Measurement}
\vspace{-10pt}
\end{figure*}

We first quantify how the reactive agents perform in the interaction scenario. In particular, we measure from three key dimensions: \textit{time}, \textit{time-to-collision (TTC)}, and \textit{lateral speed}. \textit{Time} can reveal the fluctuation during the interaction, and the other two dimensions show us more details about behavioral intensity and safety level.

To make reasonable analysis over \textit{time}, we bring in the idea of Fourier analysis, mapping the time domain to the frequency domain for an overall view of the life cycle of an interaction. In detail, we conduct the fast Fourier transform (FFT) with 512 sample rates over lateral speed on trajectories and plot the envelope of them for each model, shown in \fig{fig:Fourier Analysis}. 
Considering the area between the envelope and the frequency axis as a representation of the interaction distribution, we compare two quantitative evaluation metrics: \emph{IoU} and \emph{Coverage}. \emph{IoU} is computing the intersection over union area of model and human data, and \emph{Coverage} is defined as:
\begin{equation*}
    \text{Coverage}=\frac{\sum_{f=-N}^N v_{\text{model}}(f)}{\sum_{f=-N}^N \max(v_{\text{model}}(f),v_{\text{human}}(f))}~,
\end{equation*}
where $v(f)$ is the amplitude at frequency $f$.
For analyzing \textit{TTC} and \textit{lateral speed}, we
draw the scatter distribution at all cut-in moments in \fig{fig:Scatter Distribution}. Assuming the data distribution as Gaussian, we further show the 95\% confidence ellipse of human data and the combination of models (strategies).

In \fig{fig:Fourier Analysis}, both interactive models can cover almost all the frequency domain with high \textit{IoU} (GAIL: 0.79, InfoGAIL: 0.73) and \textit{Coverage} (GAIL: 0.91, InfoGAIL: 0.95).
For the InfoGAIL model, the shadow area indicates difference between its modalities, and such diversity are particularly evident in low and high frequency areas.
Furthermore, in \fig{fig:Scatter Distribution}, we find that the ellipse of reactive agents successfully matches most parts of human data distribution, showing its high fidelity. Additionally, the reactive agents occupy the nearby area of human distribution, which can be seen as a reasonable enhancement of interaction, showing the diversity of making rare cases.

\subsubsection{Traffic Flow Analysis}
We measure the traffic quality in a holistic view, i.e., qualify the traffic flow generated by multi-agent reactive agents.
In particular, we use affordance~\cite{Chen_2015_ICCV} as the metric, which calculates metrics with neighbors in an egocentric view. Since interaction commonly happens between ego-agent and its neighbors, this metric is amenable for analyzing interactive traffic. Here we take \emph{mean distance, speed, heading} of neighbor vehicles to describe traffic dynamically. With human data as baseline (normalized to 1), we plot statistical bar charts in \fig{fig:Traffic Flow}.

These statistics show clear evidence that the traffic around ego-agent is very similar to human data. Specifically, the max error rate in both distance and heading $\leq 7\%$, speed $\leq 12\%$.

\section{Benchmark and Optimization of Driving Strategies}
\label{sec:benchmark}
This section demonstrates two use cases of traffic flows generated by \our. First, we show that the \our traffic flow can be used to assess the performance of given driving strategies. Specifically, we demonstrate the results of using replay trajectories generated from guided traffic generation. Then, we demonstrate that the \our environment successfully optimizes policy trained on history replay traffic flows.

\subsection{Benchmark under \our Traffic Flow}
We benchmarked four popular AD solutions, including three machine learning algorithms, Behavior Cloning (BC), Generative Adversarial Imitation Learning (GAIL), Soft Actor-Critic~\cite{pmlr-v80-haarnoja18b} (SAC), and a rule-based agent using keep-lane strategy based on the intelligent driver model (IDM). 

We evaluate the safe-driving ability of algorithms under the interactive traffic flow built by \our. Differing from quality measurement, here we design strong interaction patterns and let the ego vehicle deal with constant active behavior from neighboring social vehicles that affect its normal driving. 

\begin{figure*}[ht]
    \centering
    \begin{subfigure}[b]{0.23\textwidth}
        \centering
        \includegraphics[width=\textwidth]{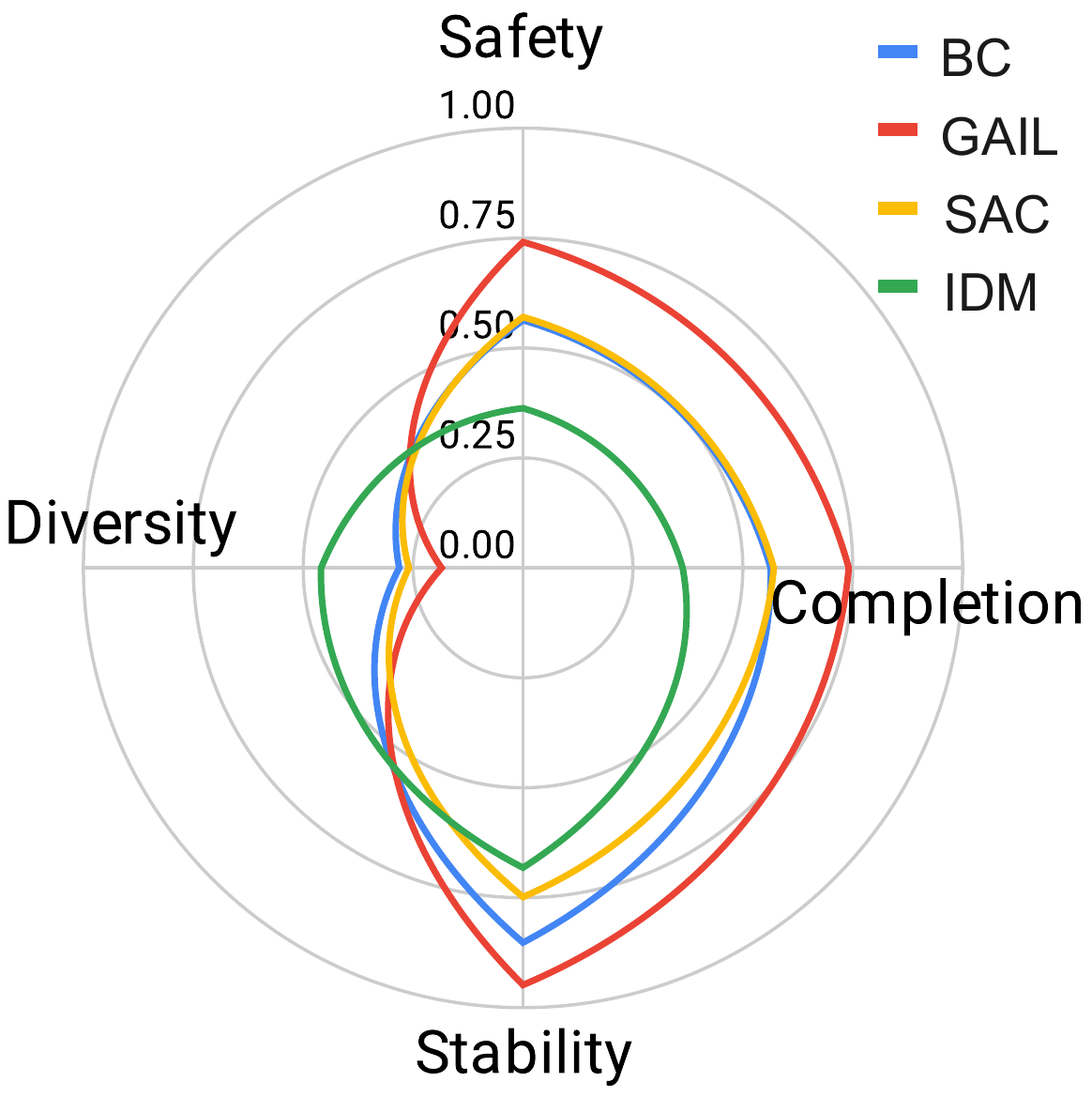}
        \caption{Left Cut-in}
    \end{subfigure}
    \begin{subfigure}[b]{0.23\textwidth}
        \centering
        \includegraphics[width=\textwidth]{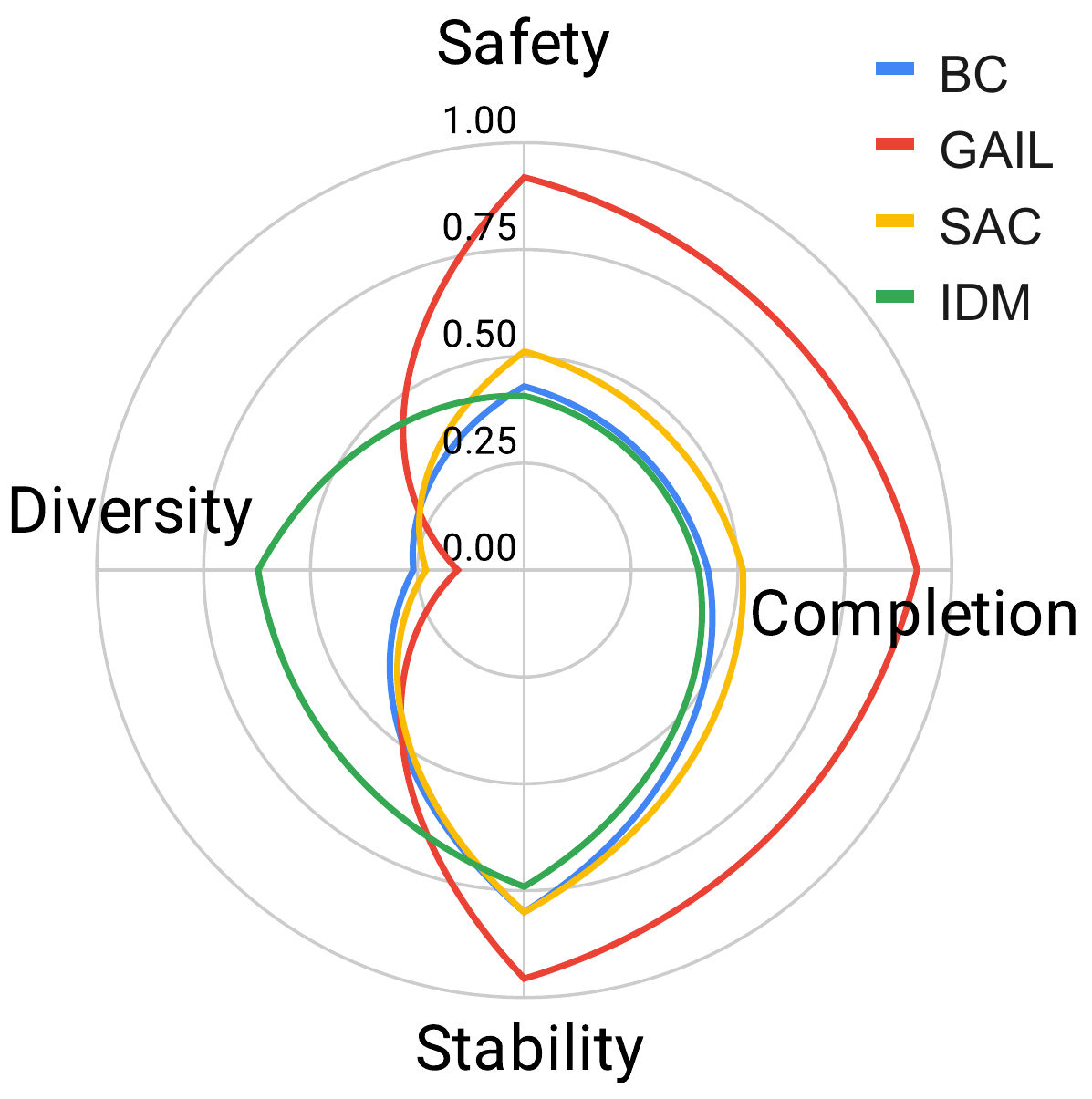}
        \caption{Right Cut-in}
     \end{subfigure}
    \begin{subfigure}[b]{0.23\textwidth}
        \centering
        \includegraphics[width=\textwidth]{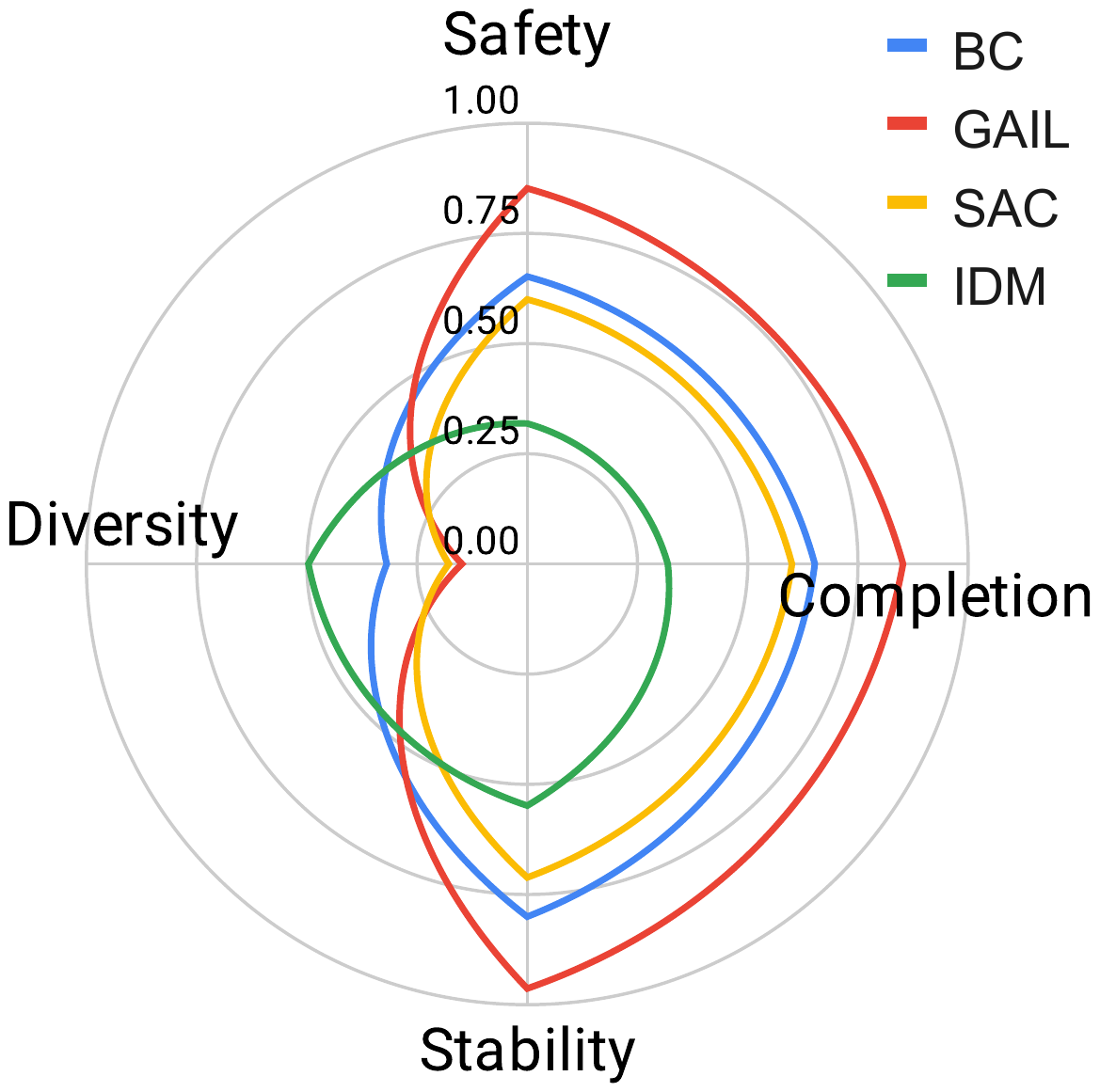}
        \caption{On-ramp}
     \end{subfigure}
    \begin{subfigure}[b]{0.23\textwidth}
        \centering
        \includegraphics[width=\textwidth]{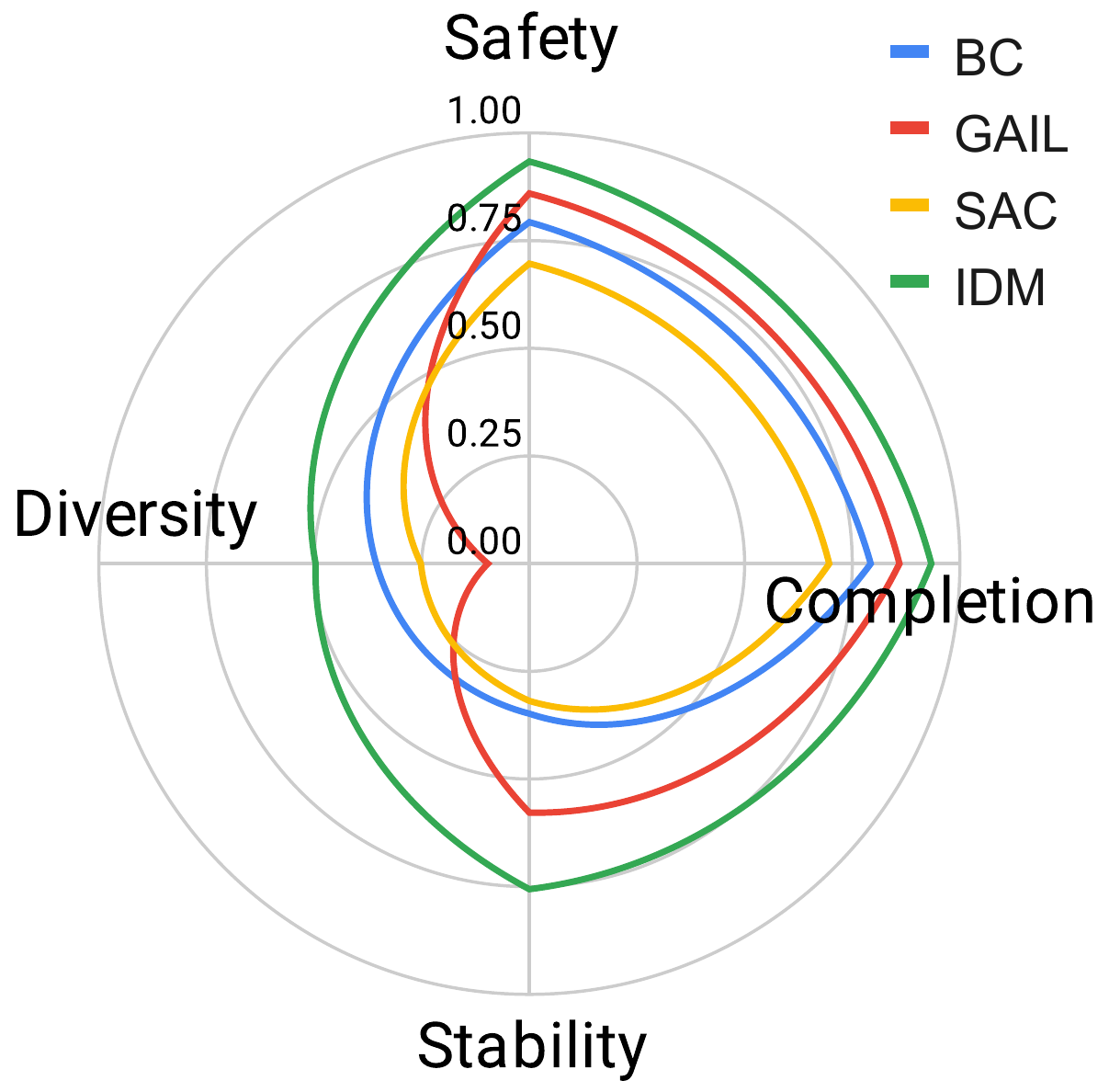}
        \caption{Off-ramp}
     \end{subfigure}
\caption{Benchmark results of ego strategies. To keep consistency, we use the same metric as \fig{fig:Model Performance}. Notice that we evaluate the safe-driving ability here, so Safety is numerically equal to Completion.}
\label{fig:Benchmark Results}
\vspace{-10pt}
\end{figure*}

The results are depicted in Fig \ref{fig:Benchmark Results}, where GAIL shows the best performance regarding safety and stability in three of the four tasks. The reason may be that GAIL can efficiently use expert training data and interact online with the simulator. On the contrary, IDM shows poor performance because of frequent interaction in the traffic, which the model itself cannot handle properly. The only exception is the off-ramp task, where GAIL underperforms IDM and ranks the second.
We argue that it is because the interactions between vehicles in the Off-ramp scenario are not complex, and rule-based policies can already achieve good results. Too much modeling of interactions between vehicles by GAIL will instead negatively affect the driving safety and stability.
Also, we observed that all algorithms perform well on fix-located traffic tasks like on-ramp and off-ramp, which may imply interactive traffic with moving bubble has a more challenging dynamic to deal with. 

\paragraph{Benchmark using Guided Generation}
For the left cut-in environment, we conduct guided sampling of replay trajectories according to driving models, and evaluation results on these trajectories are shown in \tb{table:guided}.
The definition of \textit{Stability} is the same as in \se{sec:model-performance}, and \textit{Distance Ratio} is the ratio of the actual distance traveled by the ego vehicle in the simulator to the distance traveled in the replay trajectory. The upper limit of the distance ratio is set to 1.

Specifically, we compare guided sampling (Guided) with two other sampling techniques, dataset sampling (Dataset) and normal sampling (Normal). Dataset sampling first samples a period of 128 time steps (12.8 seconds) from the dataset and randomly chooses one vehicle that appeared in this interval as the ego vehicle. Then the other 15 vehicles, which are most close to the ego vehicle at the first step, are chosen as social vehicles. Trajectories obtained by dataset sampling are also used to train the diffusion-based generative model.
Normal sampling stands for a direct sample from the generative model. Guided sampling trains the scoring function network using the model performance on the dataset sampling trajectories, and its gradient is used to guide the generative model to produce samples resulting in the models' low performance. For all three sampling methods, we sample 1000 trajectories for evaluation.

From \tb{table:guided}, we observe that ego vehicles' performance is similar when being evaluated on replay trajectories obtained by dataset and normal sampling. This shows that the trajectories sampled by the generative model are similar to the dataset trajectories in evaluating the performance of the ego strategies. On the other hand, both stability and distance ratio evaluated on the guided generated trajectories are significantly lower than the others. In particular, the distance ratio of GAIL model is lowered by 16.1\% with guided sampling, and the stability of SAC model is lowered by 10.9\%.

\begin{table}
\begin{center}
    \caption{Comparisons of different trajectory sampling techniques.}
    \label{table:guided}
    \vspace{-5pt}
    \begin{tabular}{ccccc}
    \toprule
    Metric & Sample Type & \textbf{GAIL} & \textbf{SAC} & \textbf{BC} \\
    \midrule
    \multirow{3}{*}{Stability} & Dataset & 0.911 & 0.811 & 0.822 \\
    & Normal & 0.908 & 0.807 & 0.824 \\
    & Guided ($\downarrow$) & 0.861 & 0.719 & 0.775 \\
    \midrule
    \multirow{3}{*}{Distance Ratio} & Dataset & 0.745 & 0.663 & 0.594 \\
    & Normal & 0.734 & 0.653 & 0.591 \\
    & Guided ($\downarrow$) & 0.616 & 0.559 & 0.549 \\
    \bottomrule
    \end{tabular}
\end{center}
\end{table}

\subsection{Optimization under \our Traffic Flow}

\begin{table}
\begin{center}
    \caption{Completion rates of two models before and after being fine-tuned under \our traffic flow in the cut-in scenario.}
    \vspace{-5pt}
    \label{table:2}
    \begin{tabular}{ccccc}
    \toprule
    & \textbf{GAIL} & \textbf{GAIL-finetune} & \textbf{SAC} & \textbf{SAC-finetune} \\
    \midrule
    Replay & 0.795 & 0.806 & 0.733 & 0.676 \\
    \our & 0.766 & 0.878 & 0.712 & 0.882 \\
    \bottomrule
    \end{tabular}
\end{center}
\vspace{-10pt}
\end{table}

To further claim the advantage of \our, we take the history replay environment as our baseline and conduct optimization tasks. We first train two policies using SAC and GAIL, respectively, until convergence. Then we execute fine-tuning with \our interactive traffic flow and evaluate them under the history replay environment and \our environment. 

The result can be seen in Table \ref{table:2}. It suggests two significant facts: (1) Policy trained by history replay shows relatively poor performance in the \our environment, indicating that it can hardly handle a dynamic environment with realistic human response. (2) Policy fine-tuned by \our achieves high performance in both evaluations, implying that \our can output a more robust policy than static human replay data. Given that surrounding vehicles can make reasonable responses even if the policy itself deviates from the dataset trajectory in the \our environment, it somehow means that collisions happening in the \our environment are more unacceptable. The performance drop may suggest a not small loss in interaction ability and will cause agent collision when facing similar real-world situations.

\section{RELATED WORK}
\subsection{Microscopic Traffic Simulation}

Unlike macroscopic traffic simulation, which models average vehicle dynamics like traffic density, microscopic traffic simulation separately models each vehicle and its dynamics, playing a critical role in optimizing self-driving strategies in simulators. Most traffic simulators or benchmarks use heuristic-based models to simulate the background traffic~\cite{gipps1981behavioural, treiber2000congested, elsayed2020ultra}, e.g., following the lane and avoiding head-on collisions~\cite{dosovitskiy2017carla}. 
Since human driving behaviors are hard to define completely with heuristic rules, these methods lack the ability to model complex multi-vehicle interactions. SimNet~\cite{bergamini2021simnet} is similar to our approach in using a data-driven approach to obtain models for social vehicle control. However, the control models in SimNet are trained in an offline manner, which is more sensitive to distribution shifts.

Microscopic traffic simulation solutions in autonomous driving simulators need to offer easy-to-use interfaces to allow users to specify specific characteristics of the traffic flow.
However, most traffic generation algorithms do not provide such interfaces~\cite{bhattacharyya2020modeling, bansal2018chauffeurnet}, so the additional design is required when incorporating into simulators. \our, as a complete microscopic traffic generation framework independent of the simulator, integrates the interface for traffic definition in \ourkit.

Another group of studies on microscopic traffic simulation focus on generating traffic flows that make self-driving vehicles perform poorly. Ding et al.~\cite{ding2020learning} propose to generate safety-critical scenarios by sampling from a pre-designed probability graphic model, where conditional probabilities are optimized via the policy gradient. Such optimization with reinforcement learning does not utilize real datasets, and the generated scenarios may lack fidelity. AdvSim~\cite{wang2021advsim} conducts black-box adversarial attacks on real-world trajectories by adding perturbations to vehicles' behaviors. However, random perturbations may drive the final generated samples away from the true data distribution. Although the aforementioned corner-case generation methods can be used to evaluate the worst performance of a strategy, such unrealistic scenarios violate the fidelity requirement of \our. Instead, \our provides an adversarial sampling of scenarios given the current strategy, while constraining the generated scenarios to stay within the real data distribution.

\begin{table}[h!]
\vspace{-2pt}
\caption{Comparison of microscopic traffic flows in driving simulations. Sim, bench, comp are shorts for simulator, benchmark, component respectively.}
\vspace{-4pt}
\label{tb:traffic-flow-comparison}
\centering
\resizebox{0.98\linewidth}{!}{
\begin{tabular}{ccccc}
\toprule
\multirow{2}{*}{Name} & Data-driven & Configurable & Adversarial & \multirow{2}{*}{Type} \\
& Models & Interface & Generation & \\
\midrule
CARLA \cite{dosovitskiy2017carla} & \xmark & \cmark & \xmark & Sim \\
SMARTS \cite{zhou2020smarts} & \xmark & \cmark & \xmark & Sim \\
BARK \cite{bernhard2020bark}& \cmark & \cmark & \xmark & Bench \\
NuPlan \cite{caesar2021nuplan} & \cmark & \cmark & \xmark & Bench \\
SimNet \cite{bergamini2021simnet}& \cmark & \xmark & \xmark & Comp \\
AdvSim \cite{wang2021advsim}& \xmark & \xmark & \cmark & Comp \\
\midrule
\textbf{\our (Ours)} & \cmark & \cmark & \cmark & Comp \\
\bottomrule
\end{tabular}}
\vspace{-6pt}
\end{table}

We compare microscopic traffic flows in existing driving simulation literature from three aspects in \tb{tb:traffic-flow-comparison}. Specifically, we judge whether these traffic flows are generated by data-driven driving models, provide a configurable interface for controllable generation, and support generating specific rare-case traffic.

\subsection{Human Behavior Modeling}

To more accurately evaluate the performance of the autonomous driving model in real traffic, we choose to deploy social vehicle models that imitate human drivers in the simulator. This requires us to adopt a practical approach to human behavior modeling. 
Human behavior modeling is becoming a trending research direction in the field of human-robot interaction (HRI) and has been used for various purposes. A direct motivation is to obtain control policies by imitating human data and deploying the imitated models on robots or autonomous vehicles~\cite{huang2015adaptive, codevilla2018end, bhattacharyya2020modeling}.
Other than directly using human behavior models to control the robots, these models can also help robots make decisions by predicting the actions of humans who interact with them~\cite{mainprice2016goal, schwarting2019social}. Moreover, if learned human models are conditioned on the robots' actions, they can be used to reason future human responses to robot behavior, which enables the robots to proactively shape or guide human behaviors~\cite{dragan2015effects, tellex2014asking}.
Another goal of building human behavior models is to build more realistic simulators that can better access or improve the model performances in human-robot interactions without real interaction with humans~\cite{shi2019virtual, caesar2021nuplan}. 
\our aims to accomplish this goal in autonomous driving tasks and adopts several adversarial imitation learning methods to build human behavior models. Since human behavior modeling is attracting more and more research attention, more advanced human modeling techniques can be continuously incorporated into the \our framework.

\section{Conclusions}
This paper presents \our, a framework for generating high-quality traffic flow in a driving simulator.
\our contains two main components.
\ourbk learns vehicle control and static trajectory generation models from real-world datasets. Besides, built on top of \ourbk, \ourkit provides an easy-to-use interface to customize the traffic flow. Combining these two modules, \our can deliver traffic flow with high fidelity, diversity, and controllability. Under \our, we design two benchmark tasks with highly interactive traffic flow. Under these two tasks, we conduct experiments to show the high quality of generated traffic flows from multiple perspectives. Also, we demonstrate two use cases of \our: evaluating the performance of existing driving strategies and 
fine-tuning those driving strategies. 

\our is already in the engineering schedule and will soon be integrated into SMARTS simulator. We look forward to making \our a standard component of more driving simulators.


\begin{acks}
The Shanghai Jiao Tong University team is supported by National Key R\&D Program of China (2022ZD0114804), Shanghai Municipal Science and Technology Major Project (2021SHZDZX0102) and National Natural Science Foundation of China (62076161, 62177033).
\end{acks}

\bibliographystyle{ACM-Reference-Format}
\bibliography{ref}

\clearpage
\newpage
\appendix



\section{Simulation Space}
\subsection{Observation Space}
\label{sec:appendix-obs}
To let models be generalized in different maps, we collect information from three main aspects in egocentric coordinates: [\emph{Ego dynamics, Lane observation, Neighbor observation}].
\subsubsection{Ego dynamics}
Ego dynamics choose the absolute value to represent ego vehicle attributes. Here we simply use \emph{linear velocity} since heading and position are relative amounts changing with the map.

\subsubsection{Lane observation}
In SMARTS, the simulator produces a list of equally spaced points in the centerline of lanes on the map, called waypoints. Each waypoint has its position and heading. Therefore, by calculating relative data between the nearest waypoint and the ego vehicle, we can locate our position on the map. We here calculate the relative position and heading of the ego, left, and right lane between agent and nearest waypoint.

\subsubsection{Neighbor observation}
\begin{figure}[htbp]
\centering
    \includegraphics[width=0.95\linewidth]{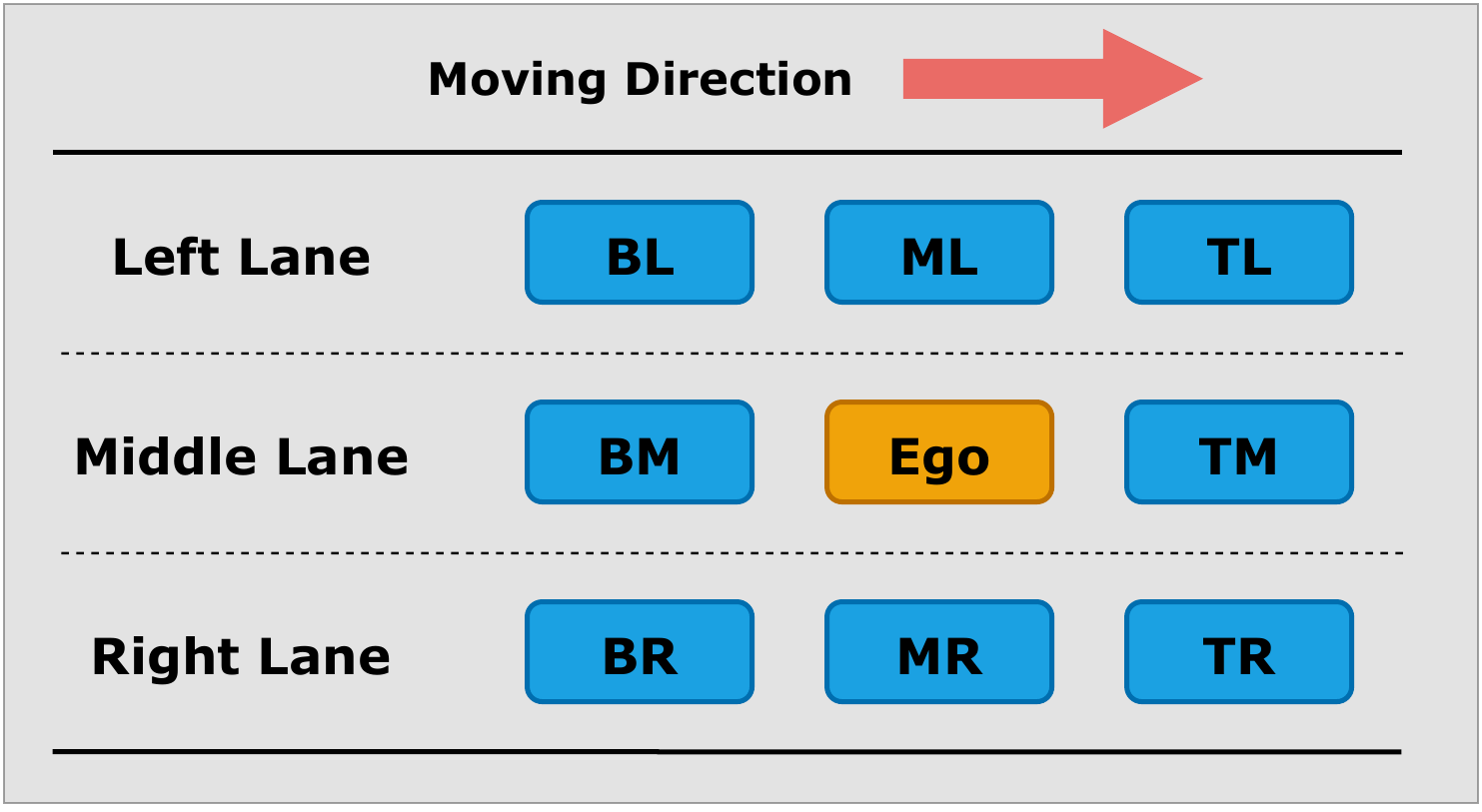}
\caption{Illustration of relative neighbor positions to the ego vehicle. The first letter indicates the neighbor's relative position, while the second indicates its lane.}
\label{fig:Neighbor_Position}
\end{figure}

We divided neighbor vehicles into eight areas according to their relative position, illustrated in Fig \ref{fig:Neighbor_Position}. The first letter indicates its position: ['B'ottom, 'M'iddle, 'T'op] while the second indicates its lane: ['L'eft, 'M'iddle, 'R'ight]. We calculate the relative position, heading, and speed for each neighbor.

\subsection{Action Space}
We here take continuous 2-dimensional \emph{linear acceleration} $a_l$ and \emph{angular velocity} $w_a$ as our action space. As SMARTS deploys a dynamic vehicle model with physical constraints, and we want to make reasonable behavior compared to the NGSIM dataset, we here make corresponding limitation for $a_l \in [-3.0,3.0] m/s^2$ and $w_a \in [-2.0,2.0] rad/s$.





\section{Model Zoo}
The tabular model zoo used for constructing interaction scenarios is shown in Table \ref{table:Model Zoo}. We have trained common reactive models as well as models dealing with specific interactions.

\begin{table*}[htbp]
\begin{center}
    \begin{tabular}{c|c|c}
    \toprule
    \textbf{Algorithm} & \textbf{Training Scenarios} & \textbf{Model} \\
    \hline
    GAIL & left cut-in, right cut-in, on-ramp, off-ramp & left cut-in, right cut-in, on-ramp, off-ramp \\
    MAGAIL & all scenarios & reactive \\
    InfoGAIL & left cut-in, right cut-in, on-ramp, off-ramp & left cut-in, right cut-in, on-ramp, off-ramp[$c$ = 1,2,3] \\
    \bottomrule
    \end{tabular}
    \vspace{5pt}
    \caption{Algorithms and corresponding scenarios we use to train the models in the model zoo. For GAIL and InfoGAIL, each interaction scenario trains a corresponding model. For MAGAIL, we get a general reactive model using the parameter-sharing technique. We can change the modality of InfoGAIL model by assigning a different value of latent variable $c$.}
    \label{table:Model Zoo}
\end{center}
\end{table*}

\section{Evaluated Tasks Illustration}
\label{sec:task-illustration}

\begin{figure*}[htbp]
  \centering
  \begin{subfigure}{0.4\textwidth}
    \centering
    \includegraphics[width=\textwidth]{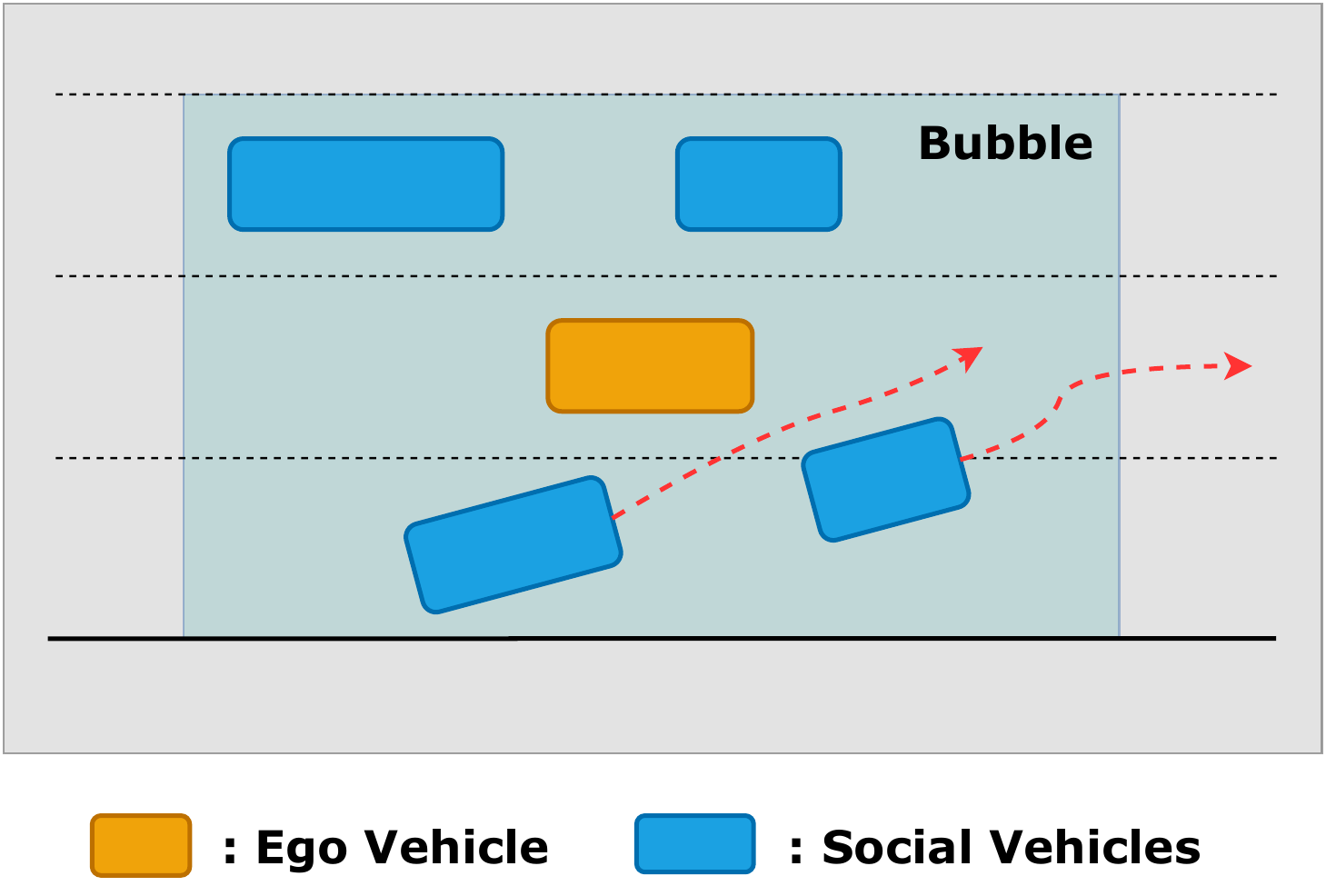}
    \caption{Left Cut-in Scenario}
  \end{subfigure}
  \hspace{10pt}
  \begin{subfigure}{0.4\textwidth}
    \centering
    \includegraphics[width=\textwidth]{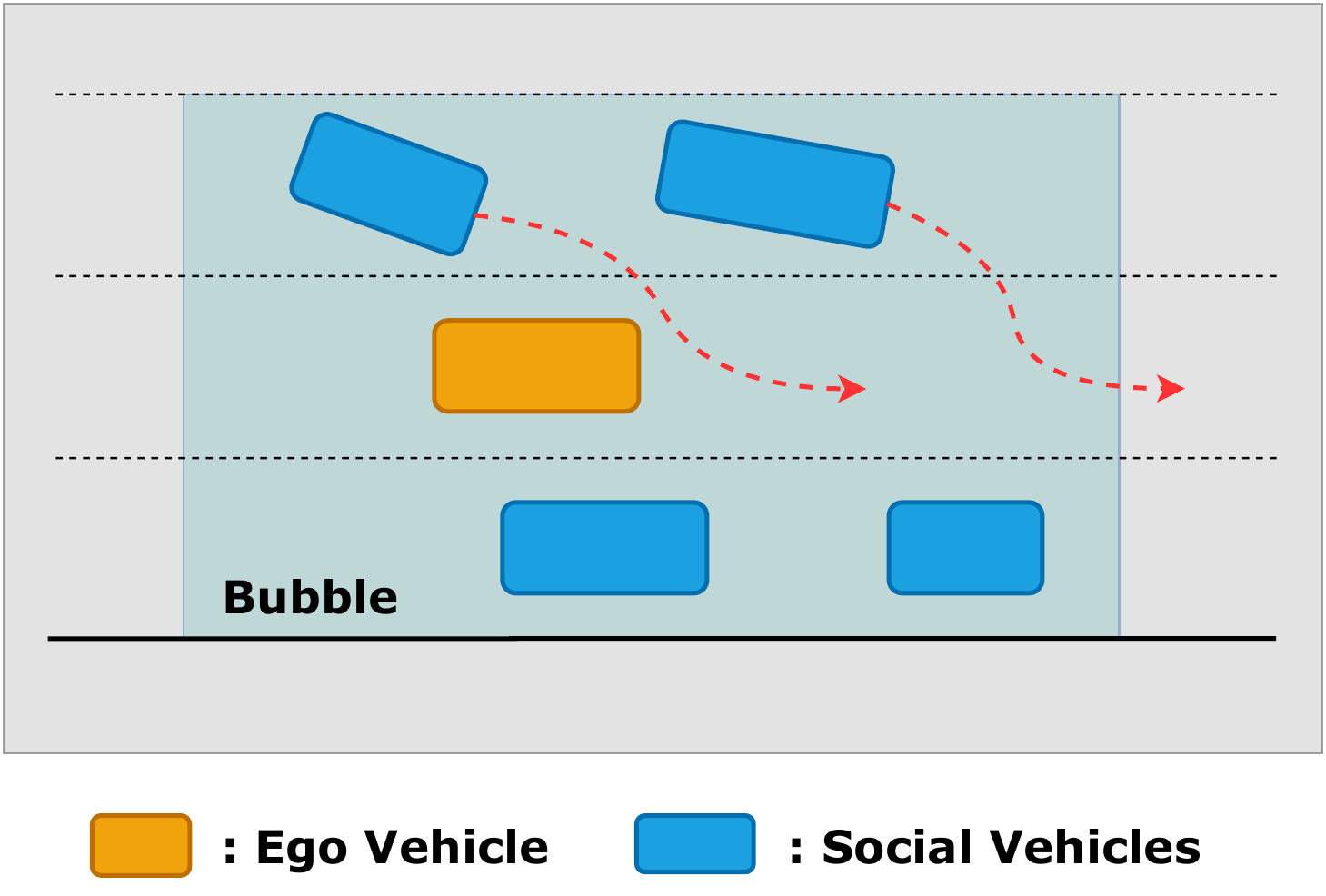}
    \caption{Right Cut-in Scenario}
  \end{subfigure}
  \vskip\baselineskip
  \begin{subfigure}{0.4\textwidth}
    \centering
    \includegraphics[width=\textwidth]{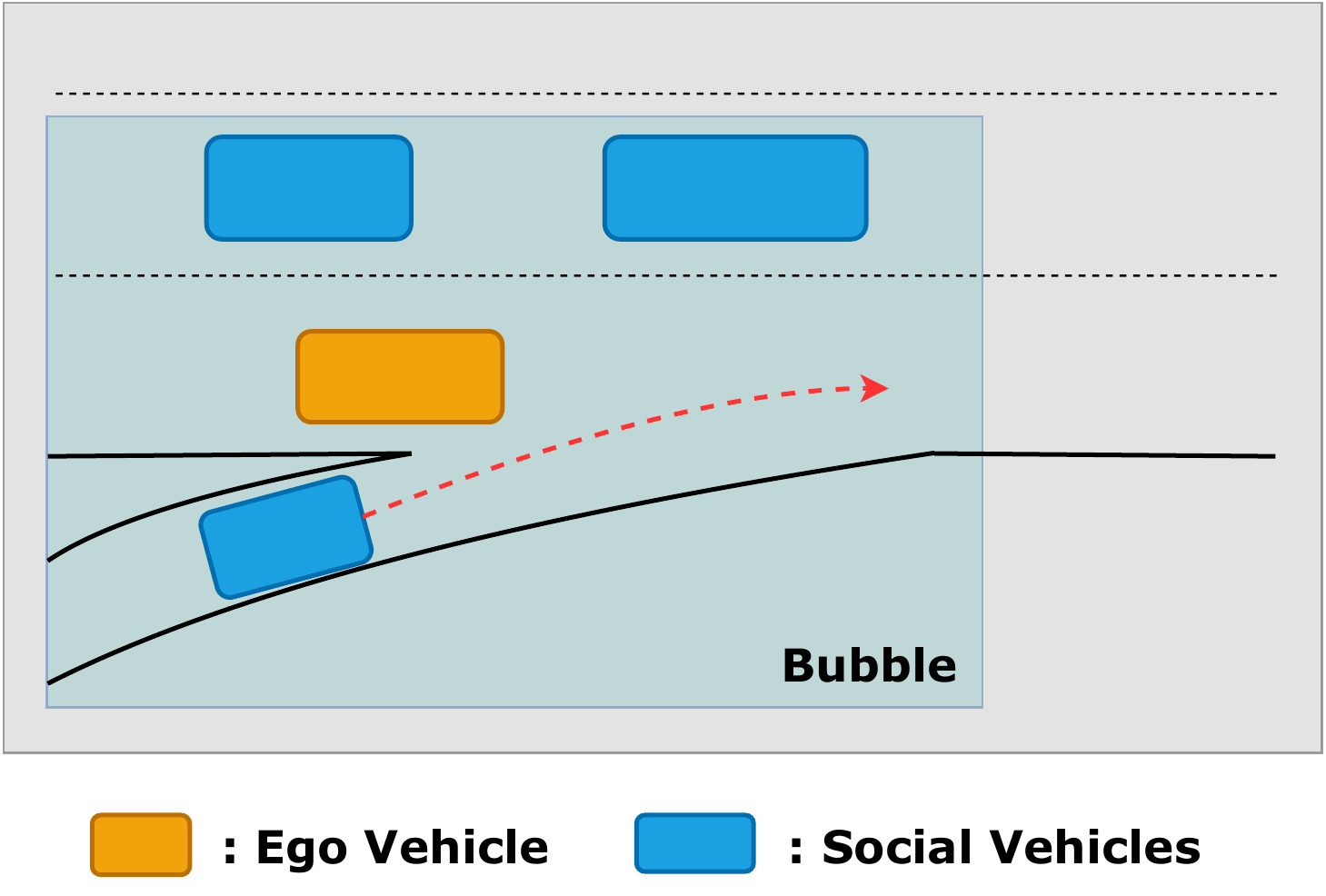}
    \caption{On-ramp Scenario}
  \end{subfigure}
  \hspace{10pt}
  \begin{subfigure}{0.4\textwidth}
    \centering
    \includegraphics[width=\textwidth]{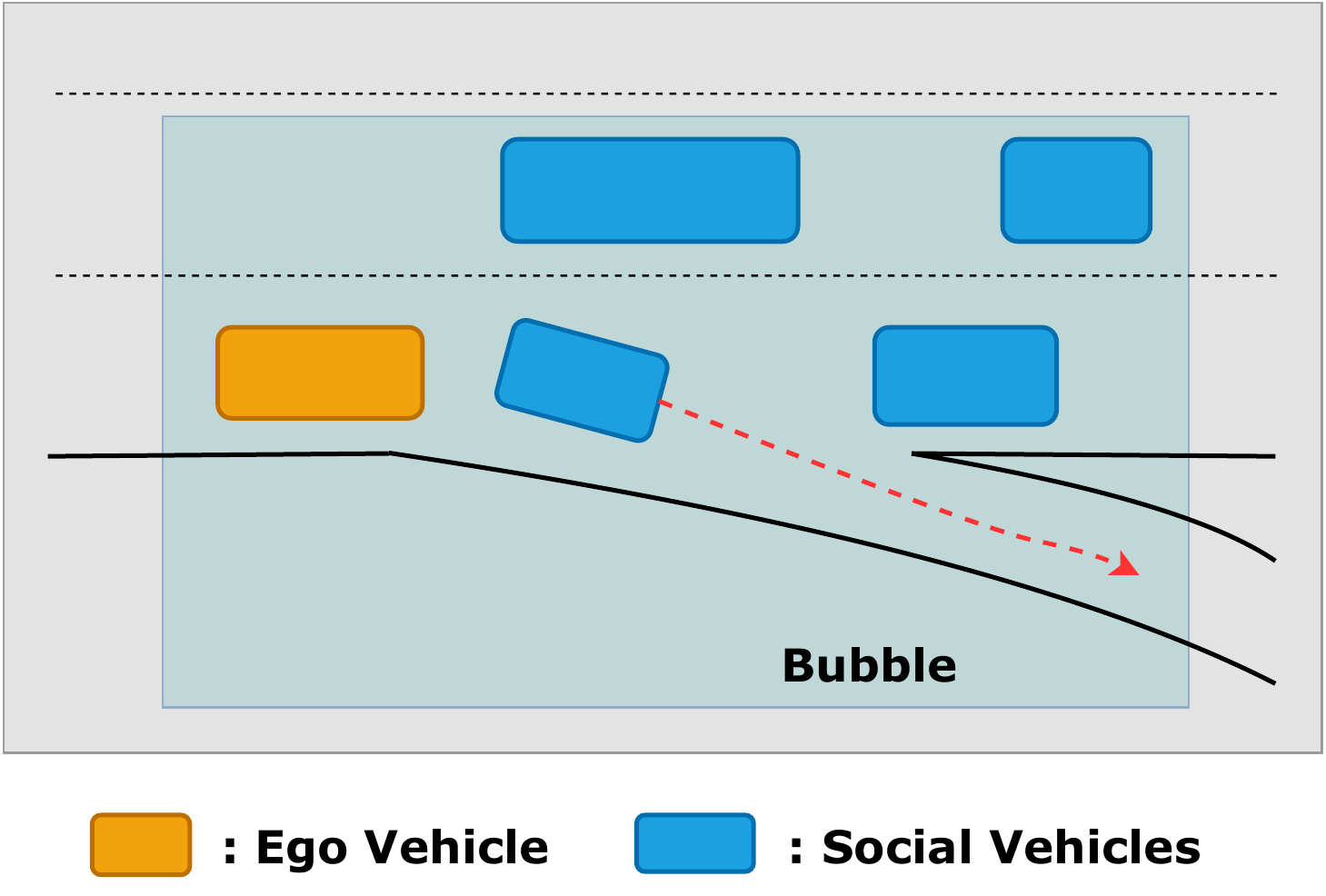}
    \caption{Off-ramp Scenario}
  \end{subfigure}
  \caption{Schematic illustration of four evaluated tasks.}
  \label{fig:illustrate-task}
\end{figure*}

For better understanding of four example tasks mentioned in \se{sec:tasks}, we provide an illustration figure for each task in \fig{fig:illustrate-task}.

\section{Scenario Generation Examples}
\label{appendix: Scenario Example}
In \fig{fig:code-example}, we give extra code examples for creating specific interactive traffic following the scheme in the main paper. While in the main paper, we give a brief introduction with built-in configuration, here we make simple reconstruction to help users understand the \our structure. It is welcome to override the interaction handler and other interfaces to formulate the wanted scenario.

\begin{figure*}[htbp]
    \centering
    \begin{minipage}{0.45\linewidth}
    \begin{Verbatim}[numbers=left, xleftmargin=8mm]
ScenarioMaker(
    scenario = "ngsim-us101",
    bubble = Bubble(
        type='moving',
        zone=Rectangle(40,20)),
    interaction = CutinHandler(
        assign_type='common',
        cutin_direction='left',
        checker='lane_change',
    ),
    guided_generation = DiffusionGen(
        guided_mode='Distance',
        target=policy,
    ),
)
\end{Verbatim}
    \caption{Left Cut-in Scenario Example}
    \end{minipage}
    \begin{minipage}{0.45\linewidth}
    \begin{Verbatim}[numbers=left, xleftmargin=8mm]
ScenarioMaker(
    scenario = "ngsim-i80",
    bubble = Bubble(
        type='fixed',
        position=(140,0)
        zone=Rectangle(110,15)),
    interaction = OnRampHandler(
        assign_type='rare',
        checker = RampChecker(
            ramp_lane='E3',
            main_lane='gneE01'
        )
    ),
    guided_generation = "None",
)
\end{Verbatim}
    \caption{On-ramp Scenario Example}
    \end{minipage}
    \caption{Code example to construct two example scenarios. A basic \textit{InteractionHandler} require two arguments: \textit{assign\_type} for interactive model assignment, \textit{checker} for interaction behavior identification. A basic \textit{DiffusionGen} needs to offer \textit{guided\_mode} (i.e. score function) to train $\mathcal{J}(x)$ and the \textit{target} model for simulation. }
    \label{fig:code-example}
\end{figure*}

\section{Algorithm Parameters}
Here, we describe basic information about benchmarked algorithm implementation. Algorithms used in benchmark and optimization experiments share the same setting.

\subsection{Rule-Based Model}
For the rule-based model in RITA, we adopt the IDM agent interface from the SMARTS built-in agent zoo. We use a keep-lane agent as we do not require the model to do active interaction behavior but to make reasonable responses under it.

\subsection{Machine Learning Model}
\begin{table*}[htbp]
\begin{center}
\begin{subtable}[t]{0.3\textwidth}
\centering
\begin{tabular}{|c|c|}
    \hline Parameter & Value \\
    \hline Network & MLP \\
    Hidden Size & 256 \\
    Hidden Layers & 3 \\
    Batch Size & 1024 \\
    Learning Rate & 0.0003 \\
    Loss Function & MSE \\
    \hline
    \end{tabular}
    \vspace{5pt}
    \caption{BC}
\end{subtable}
\hfill
\begin{subtable}[t]{0.35\textwidth}
\centering
    \begin{tabular}{|c|c|}
    \hline Parameter & Value \\
    \hline Network & MLP \\
    Policy Hidden Size & 256 \\
    Policy Hidden Layers & 3 \\
    Discriminator Hidden Size & 64 \\
    Discriminator Hidden Layer & 2 \\
    Batch Size & 256 \\
    Policy Learning Rate & 0.0003 \\
    Discriminator Learning Rate & 0.0003 \\
    Policy Trainer & SAC \\
    \hline
    \end{tabular}
    \vspace{5pt}
    \caption{GAIL}
\end{subtable}
\hfill
\begin{subtable}[t]{0.33\textwidth}
\centering
    \begin{tabular}{|c|c|}
    \hline Parameter & Value \\
    \hline Network & MLP \\
    Policy Hidden Size & 256 \\
    Policy Hidden Layers & 3 \\
    Batch Size & 256 \\
    Alpha & 0.2 \\
    Policy Learning Rate & 0.0003 \\
    Q Learning Rate & 0.0003 \\
    Reward & Travel Distance \\
    \hline
\end{tabular}
\vspace{5pt}
\caption{SAC}
\end{subtable}
\caption{Core parameters of machine learning algorithms.}
\label{table:Algorithm Parameters}
\end{center}
\end{table*}

To conduct as fair a comparison as possible, we make all the algorithms have the same state-action space and policy network structure. We choose five random seeds for each algorithm and record their average performance. We share necessary algorithm parameters in Table \ref{table:Algorithm Parameters}.

\begin{figure*}[htbp]
\centering
\begin{subfigure}[b]{0.33\textwidth}
        \centering
        \includegraphics[width=\textwidth]{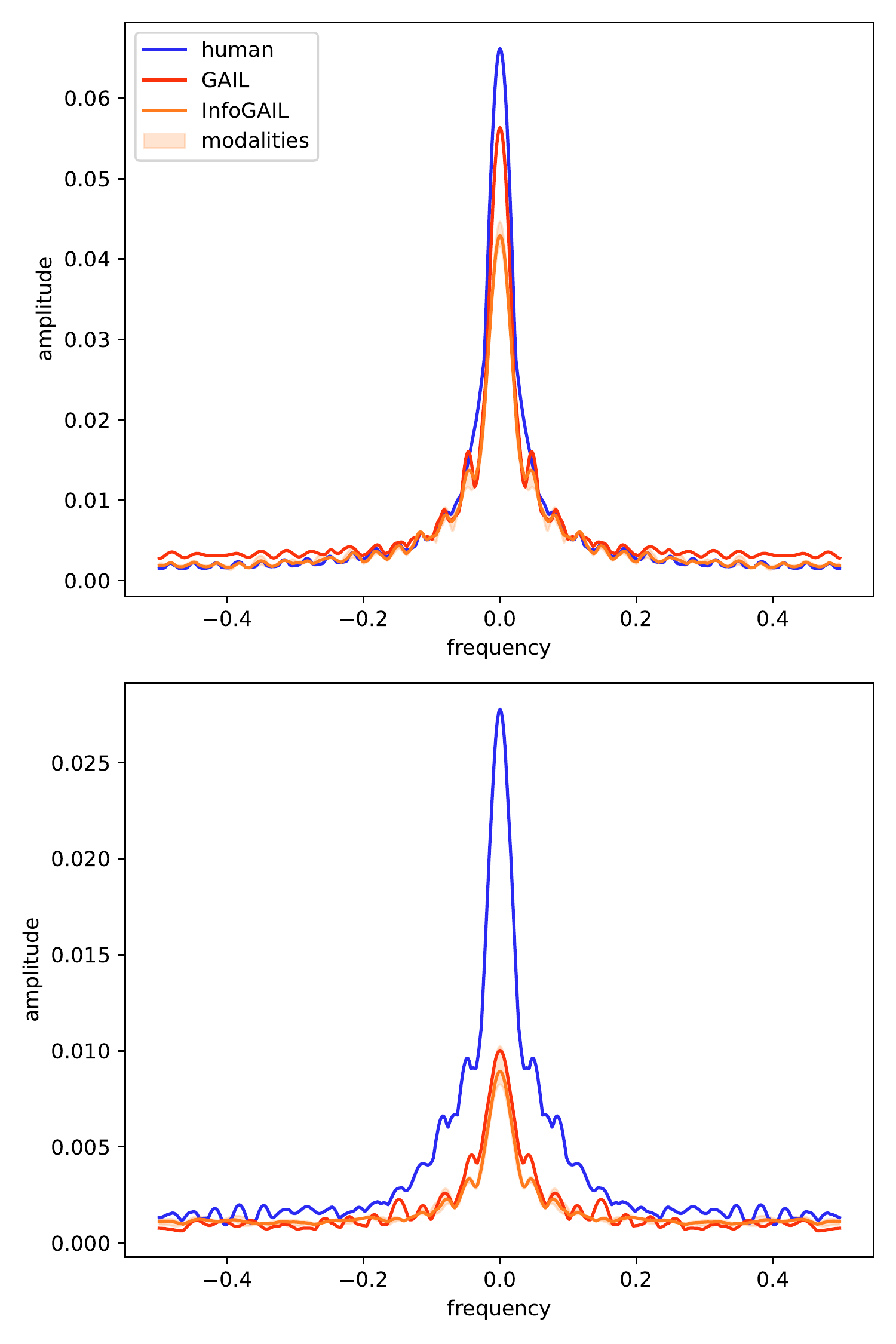}
        \caption{Fourier Analysis}
        \label{fig:Additional Fourier Analysis}
\end{subfigure}
\begin{subfigure}[b]{0.33\textwidth}
        \centering
        \includegraphics[width=\textwidth]{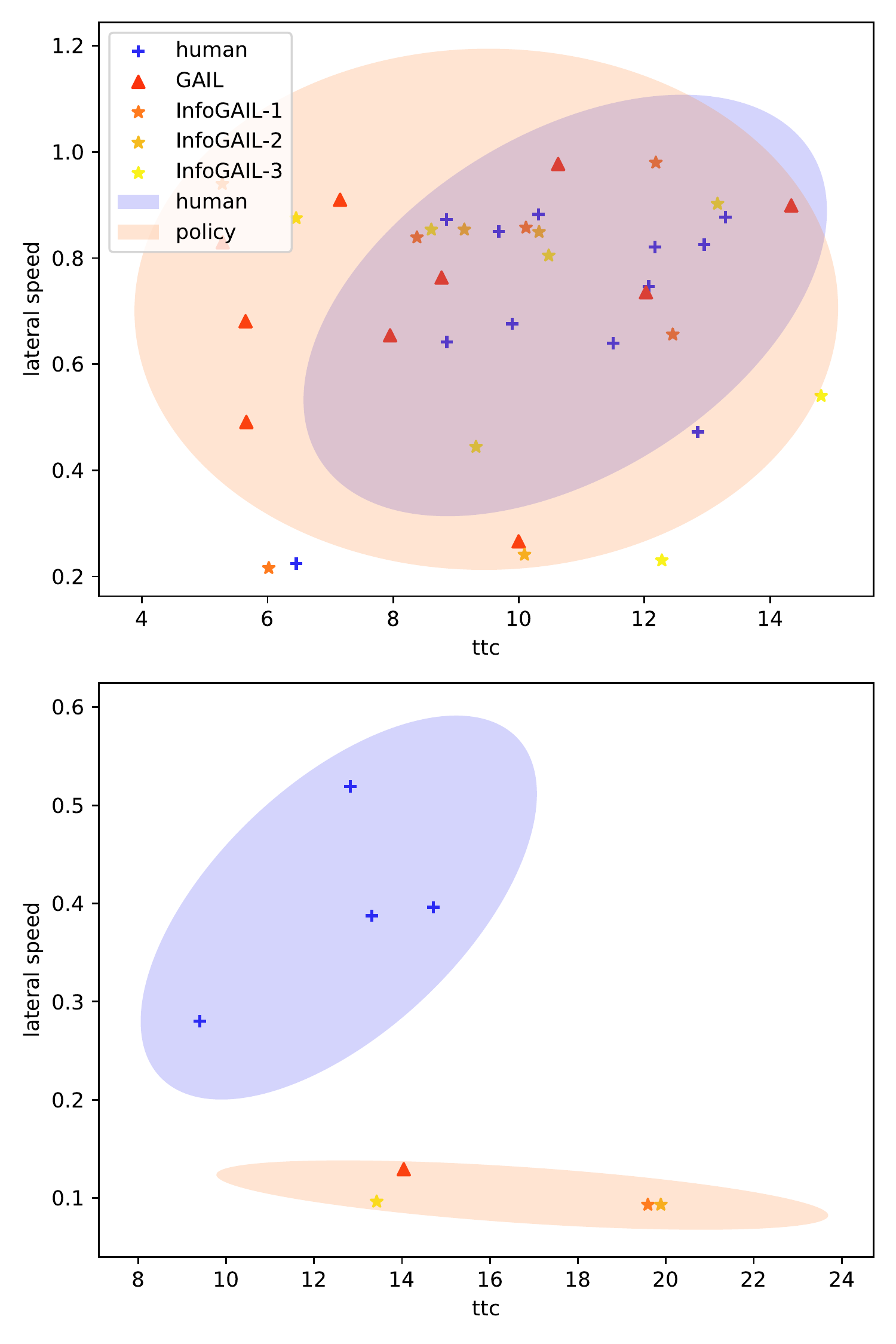}
        \caption{Scatter Distribution}
        \label{fig:Additional Scatter Distribution}
\end{subfigure}
\begin{subfigure}[b]{0.264\textwidth}
        \centering
        \includegraphics[width=\textwidth]{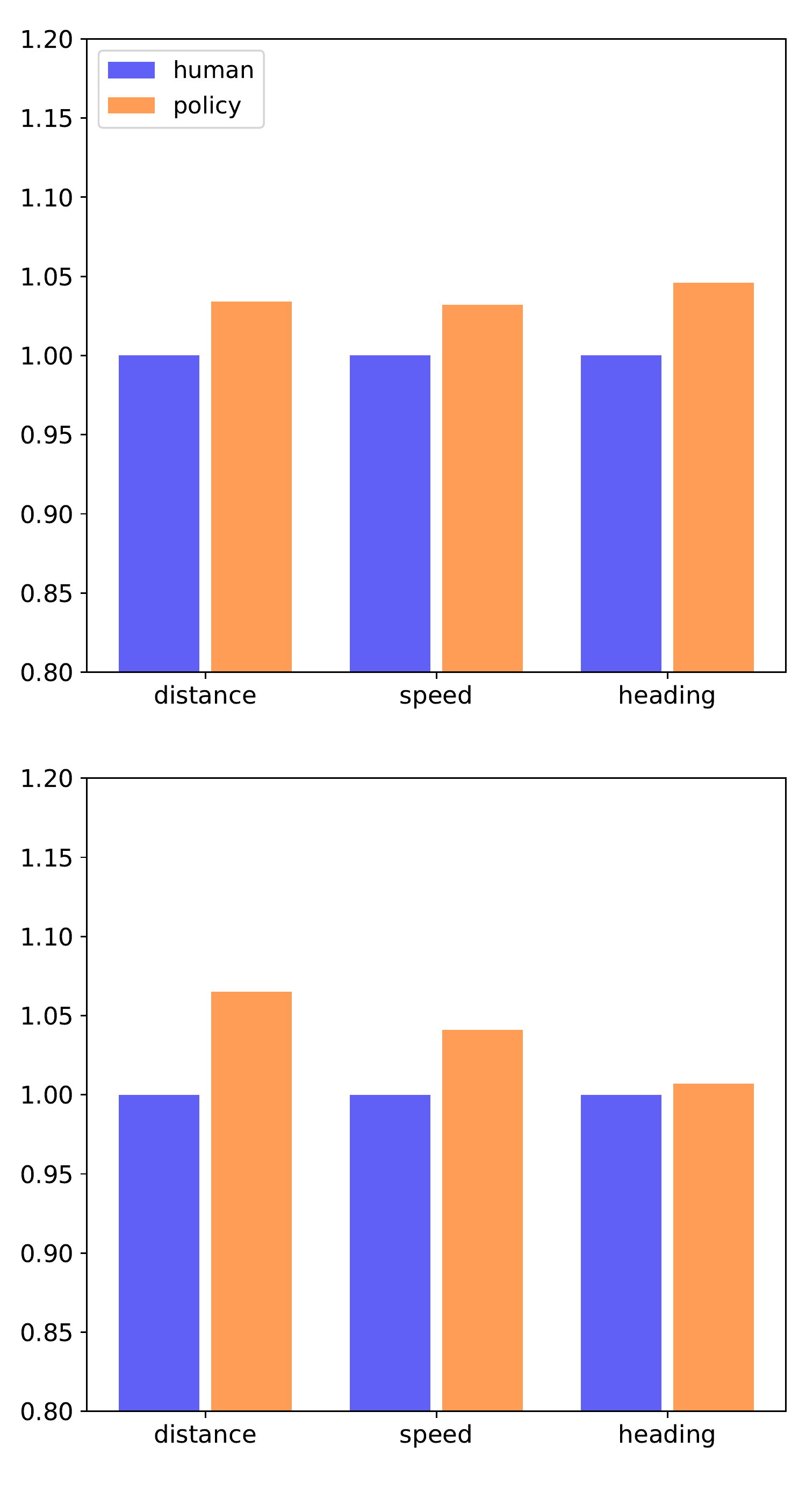}
        \caption{Traffic Flow Metrics}
        \label{fig:Additional Traffic Flow}
\end{subfigure}
\caption{Measurements of traffic flow qualities. The first row corresponds to the right cut-in scenario, and the second to the off-ramp scenario.}
\label{fig:Benchmark Quality Measurement: Appendix}
\end{figure*}

\section{Additional Results}
\label{sec:add-results}
Here we put the interaction traffic analysis on right cut-in scenario and off-ramp scenario. To achieve high fidelity, the data-driven model zoo must access adequate qualified interaction data to learn good policy, which means the lack of desired data harms the performance of the model and the simulated scenario. The training data of the right cut-in scenario only accounts for 1/6 of the total cut-in data. And for the off-ramp scenario, about 95\% of data follows the lane and not producing meaningful interaction, making it hard to create high-quality traffic flow. 

The interaction quality suffers from the same declination in model performance, which can be seen in Fig \ref{fig:Additional Fourier Analysis} and Fig \ref{fig:Additional Scatter Distribution},  where the average \textit{IoU} and \textit{Coverage} decreased, and scatter distribution matches poorly with human data. However, the traffic flow quality shown in Fig \ref{fig:Additional Traffic Flow} remains stable as the reactive ability not be hurt.




\end{document}